\documentclass[acmtog,authorversion]{acmart}

\usepackage{bm}
\usepackage{bbm}

\usepackage{graphicx}
\usepackage{tabularx}
\usepackage{makecell}
\usepackage{subcaption}
\usepackage{multirow}
\usepackage{amsmath}
\usepackage{mathrsfs}

\newcommand{\vect}[1]{\bm{#1}}
\newcommand{\tildevect}[1]{\vect{\tilde{#1}}}
\newcommand{\hatvect}[1]{\vect{\hat{#1}}}

\newcommand{\eqword}[1]{{\text{#1}}}

\usepackage{xspace}

\newcommand{\keep}[1]{}
\newcommand{\old}[1]{}

\newcommand{\norm}[1]{\lVert #1 \rVert}

\usepackage{url}
\usepackage{hyperref}
\newcommand{\fig}{Figure{}~}
\newcommand{\eqn}{Equation{}~}

\AtBeginDocument{%
  \providecommand\BibTeX{{%
    \normalfont B\kern-0.5em{\scshape i\kern-0.25em b}\kern-0.8em\TeX}}}

\setcopyright{acmlicensed}
\acmJournal{TOG}
\acmYear{2023} 
\acmVolume{42} 
\acmNumber{4} 
\acmArticle{} 
\acmMonth{8} 
\acmPrice{15.00}
\acmDOI{10.1145/3592097}

\acmSubmissionID{0}

\begin{document}

\title{GestureDiffuCLIP: Gesture Diffusion Model with CLIP Latents}

\author{Tenglong Ao}
\email{aubrey.tenglong.ao@gmail.com}
\affiliation{%
  \institution{Peking University}
  \streetaddress{No.5 Yiheyuan Road, Haidian District}
  \city{Beijing}
  \state{Beijing}
  \country{China}
  \postcode{100871}
}

\author{Zeyi Zhang}
\email{carpesu@stu.pku.edu.cn}
\affiliation{%
  \institution{Peking University}
  \streetaddress{No.5 Yiheyuan Road, Haidian District}
  \city{Beijing}
  \state{Beijing}
  \country{China}
  \postcode{100871}
}

\author{Libin Liu}
\authornote{corresponding author}
\email{libin.liu@pku.edu.cn}
\affiliation{%
  \institution{Peking University \& National Key Lab of General AI}
  \streetaddress{No.5 Yiheyuan Road, Haidian District}
  \city{Beijing}
  \state{Beijing}
  \country{China}
  \postcode{100871}
}

\renewcommand{\shortauthors}{Ao, Zhang, and Liu}

\begin{abstract}
    The automatic generation of stylized co-speech gestures has recently received increasing attention. Previous systems typically allow style control via predefined text labels or example motion clips, which are often not flexible enough to convey user intent accurately. In this work, we present GestureDiffuCLIP, a neural network framework for synthesizing realistic, stylized co-speech gestures with flexible style control. We leverage the power of the large-scale Contrastive-Language-Image-Pre-training (CLIP) model and present a novel CLIP-guided mechanism that extracts efficient style representations from multiple input modalities, such as a piece of text, an example motion clip, or a video. Our system learns a latent diffusion model to generate high-quality gestures and infuses the CLIP representations of style into the generator via an adaptive instance normalization (AdaIN) layer. We further devise a gesture-transcript alignment mechanism that ensures a semantically correct gesture generation based on contrastive learning. Our system can also be extended to allow fine-grained style control of individual body parts. We demonstrate an extensive set of examples showing the flexibility and generalizability of our model to a variety of style descriptions. In a user study, we show that our system outperforms the state-of-the-art approaches regarding human likeness, appropriateness, and style correctness.
\end{abstract}



\begin{teaserfigure}
  \centering
  \includegraphics[width=0.9\textwidth]{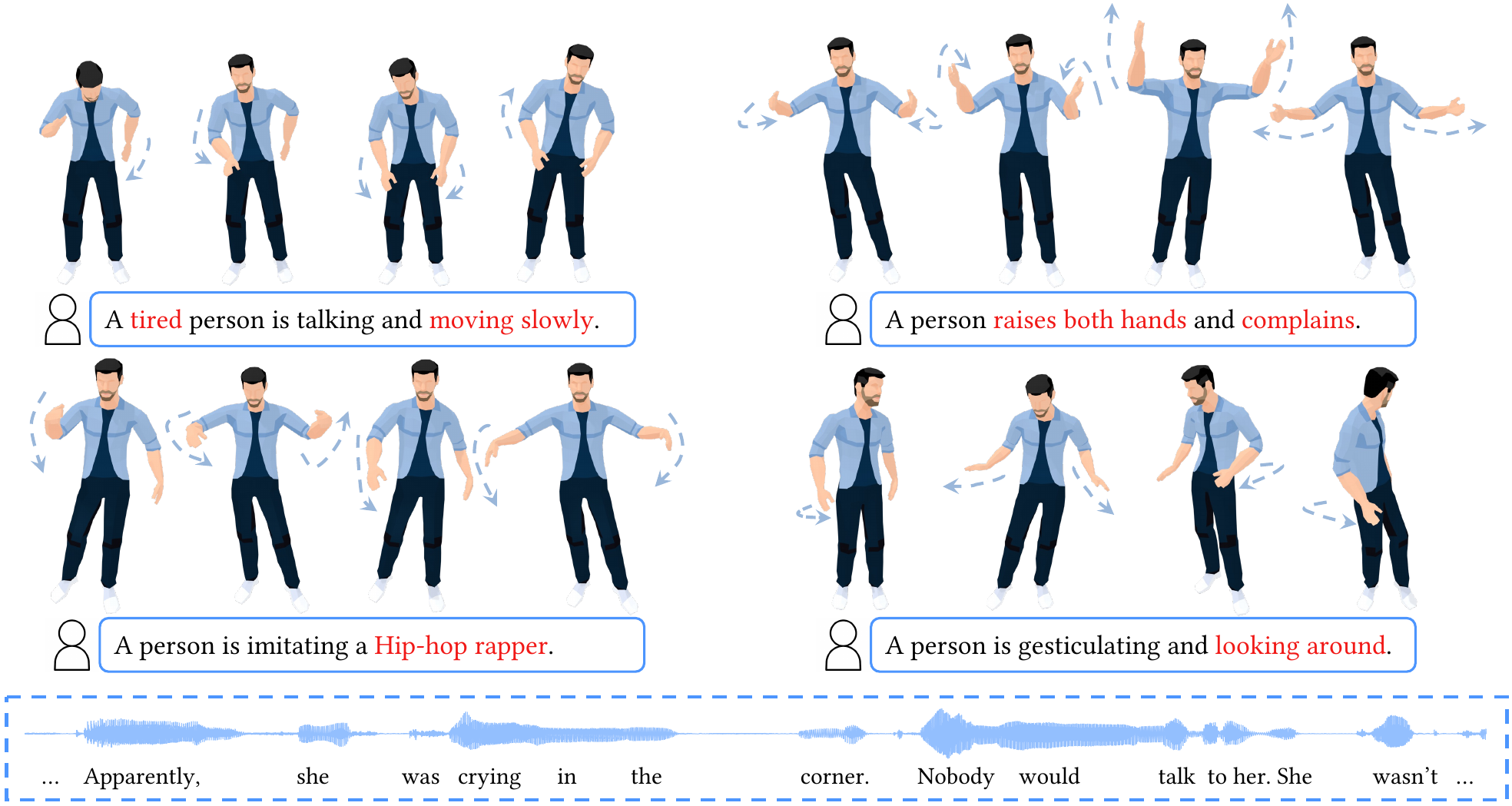}
  \caption{Stylized gestures synthesized by our system for the same speech clip conditioned on four different text prompts.}
  \Description{}
  \label{fig:teaser}
\end{teaserfigure}

\maketitle

\section{Introduction}
\label{sec:introduction}
Gestures are the spontaneous and stylized movements of hands and arms that occur while people talk. They energize the speech and reveal the idiosyncratic imagery of thoughts \cite{mcneill1992hand}. Recently, deep neural networks have been successfully applied to synthesize natural-looking gestures based on speech input, which facilitates the creation of human-like 3D avatars. However, the deep learning-based system often suffers from a lack of controllability, making synthesizing arbitrary stylized gestures under user control remain a challenging task. 

Previous neural network systems that achieve style control in gesture creation can be grouped into two categories: label-based and example-based systems. The label-based systems are typically trained on motion data with paired style labels. They allow editing of predefined styles, such as speaker identities \cite{ahuja2020stylegesture,yoon2020trimodalgesture}, emotions \cite{liu2021beatdataset}, and fine-grained styles like specific hand positions \cite{alexanderson2020stylegesture}.  However, the capacity of such systems is limited by the number and granularity of the style labels, while obtaining such style labels is costly. The example-based systems, in contrast, generate gestures of an arbitrary style by imitating an example given as a motion clip \cite{ghorbani2022zeroeggs} or a video \cite{liu2022hierarchicalgesture}. The styles characterized by these examples are often vague and can hardly convey user intent accurately. A user may need to try several times with different example data to get the  desired result.

Recently, the Contrastive-Language-Image-Pretraining (CLIP) model \cite{radford2021clip} successfully learns the connection between natural language and images. It enables several image generation systems \cite{patashnik2021styleclip,gal2022stylegannada,ramesh2022dalle2,rombach2022latentdiffusion} and motion generation systems \cite{tevet2022motionclip,zhang2022motiondiffuse,tevet2022humanmotiondiffusion} to allow users to specify desired content and style in natural language, namely, using \emph{text prompt}s. The core of the CLIP model is a large-scale shared latent space for visual and textual modalities. It learns such a space using contrastive learning, which technique can be adapted to include other modalities, such as human motions \cite{tevet2022motionclip}, into the same space. From another perspective, the CLIP model offers a flexible interface that allows users to describe their requirements accurately using multiple forms of input, such as a \emph{text prompt}, a \emph{motion prompt}, or even a \emph{video prompt}. The CLIP model can extract semantically consistent latent representation of these prompts, which can be used by an powerful generative model, such as diffusion models \cite{ho2020ddpm,song2020improvedscore}, to fulfill the user needs.

In this work, we present GestureDiffuCLIP, a co-speech gesture synthesis system that takes advantage of the CLIP latents to enable automatic creation of stylized gestures based on various style prompts. Our system learns a latent diffusion model \cite{rombach2022latentdiffusion} as the gesture generator and incorporates the CLIP representation of style into it via an adaptive instance normalization (AdaIN) \cite{huang2017adain} mechanism. The system accepts text, motion, and video prompts as style descriptors and create high-quality, realistic, semantically correct co-speech gestures. It can be further extended to allow fine-grained style control of individual body parts. 

Predicting gestures from utterances is inherently an many-to-many mapping problem. A variety of gestures can correspond to the same speech while semantic gestures and their corresponding speech are usually not perfectly aligned temporally. Such ambiguities can cause an end-to-end deep learning system to learn a \emph{mean} gesture motion and lose the semantic correspondence \cite{ao2022rhythmicgesticulator,abzaliev2022semanticclip,kucherenko2021speech2properties2gestures}.
To alleviate this problem, we learn a gesture-transcript joint embedding space using contrastive learning combined with a temporal aggregation mechanism. This joint embedding space provides semantic cues for the generator and a semantic loss that effectively guides the system to learn the semantic correspondence between gestures and speech.

Collecting a large-scale gesture dataset with diverse styles and rich, fine-grained labels is challenging. To circumvent this issue, we develop a self-supervised learning scheme to distill knowledge from the pretrained CLIP models. Specifically, we treat each gesture motion as its own style prompt and let the system reconstruct the motion based on the extracted CLIP latents. Despite never encountering other forms of style prompts during training, the system still creates satisfactory styles corresponding to arbitrary text or video prompts in a zero-shot manner.

In summary, the principal contributions of this work include:
\begin{itemize}
    \item We present a novel CLIP-guided prompt-conditioned co-speech gesture synthesis system that generates realistic, stylized gestures. To the best of our knowledge, it is the first system that supports using multimodal prompts to control the style in cross-modality motion synthesis.
    
    \item We demonstrate a successful adaptation of latent diffusion models to allow high-quality motion synthesis and propose an efficient network architecture based on transformer and AdaIN layers to incorporates style guidance into the diffusion model.
    
    \item We propose a contrastive learning strategy to learn the semantic correspondence between gestures and transcripts. The learned joint embeddings enable synthesizing gestures with convincing semantics.
    
    \item We develop a self-supervised training mechanism that effectively distills knowledge from a large-scale multi-modality pretrained model, alleviating the need for training data with detailed labels.
\end{itemize}

\section{Related Work}
\label{sec:related_work}

\subsection{Co-Speech Gesture Synthesis}
The early approaches for generating co-speech gestures often involve creating linguistic rules to translate speech input into a sequence of pre-collected gesture segments, which are typically referred to as rule-based methods \cite{cassell1994rulefullbody,cassell2001beat,kipp2004gesture,kopp2006bml}. \citet{wagner2014rulereview} provide a comprehensive review of these methods. Rule-based methods produce interpretable and controllable results, but creating gesture datasets and rules requires significant effort. To alleviate the manual effort of designing rules in rule-based methods, data-driven approaches have gradually become predominant in this field. \citet{nyatsanga2023data_driven_gesture_survey} offer a thorough survey of these methods. Early data-driven approaches aim to directly learn mapping rules from data through statistical models \cite{neff2008videogesture,levine2009prosodygesture,levine2010gesturecontroller} and combine them with predefined gesture units for gesture generation. Later, the powerful modeling capability of deep neural networks makes it possible to train complex end-to-end models using raw speech-gesture data directly. One option is deterministic models, such as MLP \cite{kucherenko2020gesticulator}, CNN \cite{habibie2021videogesture}, RNN \cite{yoon2019robot,yoon2020trimodalgesture,bhattacharya2021affectivegesture,liu2022hierarchicalgesture}, and Transformer \cite{bhattacharya2021text2gestures}. Another choice is generative models, including flow-based models \cite{alexanderson2020stylegesture,ye2022styleflowgesture}, VAEs \cite{li2021audio2gesture,ghorbani2022zeroeggs}, and VQ-VAE \cite{yi2022talkshow,yazdian2022gesture2vec,liu2022vqgesturevideo}. Due to the inherent many-to-many relationship between speech and gesture, end-to-end models can generate natural-looking gestures but face challenges in ensuring content matching between speech and generated gestures \cite{yoon2022genea}. To address this issue, some neural systems aim to explicitly model both rhythm and semantics from the perspective of model structure \cite{kucherenko2021speech2properties2gestures,ao2022rhythmicgesticulator,liu2022disco} or training supervision strategy \cite{liang2022seeg}. Furthermore, hybrid systems, such as the combination of deep features and motion graphs \cite{zhou2022gesturemaster}, have been proposed to harness the advantages of different approaches. Recently, diffusion models \cite{sohldickstein2015diffusion,song2020improvedscore,ho2020ddpm} have demonstrated impressive results in image creation \cite{ramesh2022dalle2}, human motion generation \cite{tevet2022humanmotiondiffusion, zhang2022motiondiffuse}, and gesture synthesis \cite{alexanderson2022diffusiongesture,zhang2023diffmotion}. Inspired by these works, our system adapts the latent diffusion model \cite{rombach2022latentdiffusion} for the co-speech gesture generation task and achieves appealing results.

\subsection{Style Control for Human Motion}
A typical approach to style control for human motion involves specifying a motion clip as a reference and transferring the reference clip's style to the source motion. This task is also known as \emph{style transfer}. Early works in motion style transfer integrate traditional machine learning techniques with manually defined features to infer motion styles \cite{hsu2005motion_style_translation,ma2010motion_style_transfer,xia2015realtime_motion_style_transfer,yumer2016spectral_motion_style_transfer}. Recently, deep learning-based methods have significantly enhanced motion quality. \citet{holden2016deepmotion} first propose a learning framework enabling motion style control through optimization in the motion manifold space. \citet{du2019stylemotioncvae} improve transfer efficiency by training a conditional VAE. \citet{mason2018few-shot_motion_style_transfer} use few-shot learning to generate stylized locomotion. \citet{aberman2020adain} employ a temporally invariant adaptive instance normalization (AdaIN) layer for target style injection, eliminating the need for paired data during training. \citet{wen2021stylemotionflow} achieve unsupervised style transfer using a flow model. \citet{jang2022motionpuzzle} introduce a method capable of controlling styles for individual body parts.

Previous co-speech gesture synthesis systems featuring style control can be categorized according to their dependence on style labels. Early methods requiring labeled data are limited to learning a single style for each generator. \cite{levine2010gesturecontroller,neff2008videogesture,ginosar2019stylegesture}. \citet{ahuja2022lowresource} propose a strategy that efficiently adapts a source generator to a different speaker style using low-resource data. Some works learn a speaker style embedding space with labeled speaker-motion data, enabling gesture style control by sampling from this space \cite{ahuja2020stylegesture,yoon2020trimodalgesture,bhattacharya2021affectivegesture}. \citet{alexanderson2020stylegesture} aim at controlling fine-grained styles, such as gesturing speed and spatial scope, using preprocessed control signal-motion data. Their later work \cite{alexanderson2022diffusiongesture} utilizes a diffusion model for audio-driven motion synthesis, achieving label-based style control by training the model on labeled data. For methods not requiring style labels, \citet{habibie2022motionmatching} propose a motion matching framework to achieve flexible style control. Other studies achieve arbitrary style control by imitating an example given as a video \cite{liu2022hierarchicalgesture} or a motion clip \cite{ghorbani2022zeroeggs,ye2022styleflowgesture}.  In this work, we utilize a CLIP-based encoder to extract a style embedding from an arbitrary text prompt and incorporate it into the generator via an AdaIN layer, guiding the synthesis of stylized gestures. Our system supports fine-grained multimodal style prompts as opposed to label-based style control. It employs a self-supervised learning scheme and eliminates the need for labeled data. Additionally, we use an autoregressive model rather than a parallel model, making it potentially suitable for real-time applications.
\section{System Overview}
\label{sec:system_overview}

\begin{figure}[t]
    \centering
    \includegraphics[width=\linewidth]{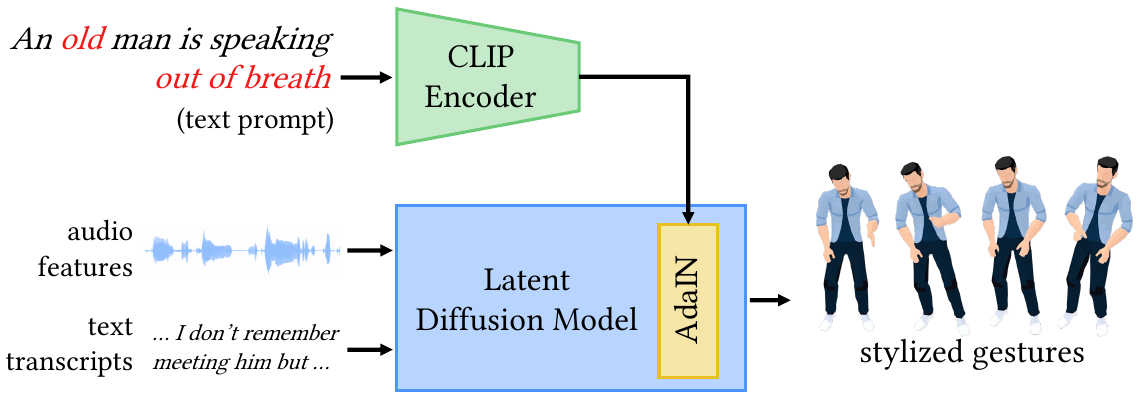}
    \caption{Our system consists of two core components: (a) a latent diffusion model that takes speech audio and transcript as input and generate co-speech gestures, and (b) a CLIP-based encoder that extracts style embeddings from an arbitrary style prompt and incorporates them into the diffusion model via an adaptive instance normalization (AdaIN) layer. The system allows using short texts, video clips, and motion sequences to define gesture styles by encoding them into the same CLIP embedding space using corresponding pretrained encoders.
    }
    \Description{}
    \label{fig:system_overview}
\end{figure}

\begin{figure*}[t]
    \centering
    \includegraphics[width=0.9\textwidth]{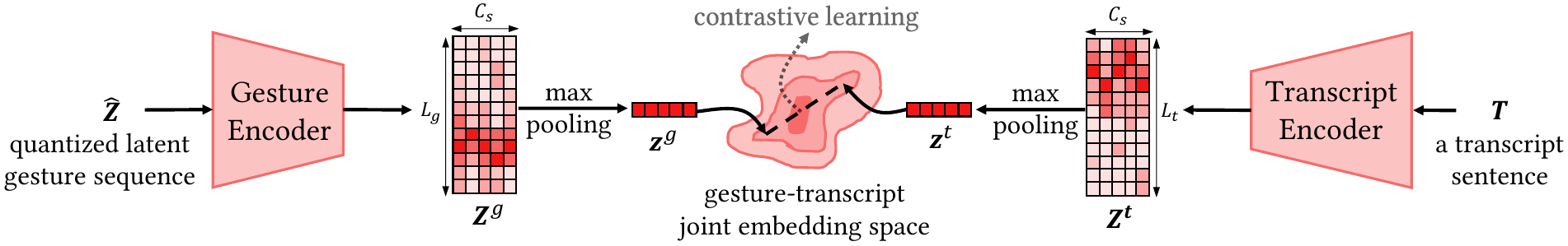}
    \caption{
        We learn an gesture-transcript joint embedding space using contrastive learning. A transcript encoder is trained to convert a transcript sentence $\vect{T}$ into a sequence of feature codes $\vect{Z}^t$, which are then aggregated into a transcript embedding vector $\vect{z}^t$ via max pooling. Similarly, the corresponding gesture sequence $\hatvect{Z}$ is processed by a gesture encoder, resulting in a feature sequence $\vect{Z}^g$ and the corresponding embedding $\vect{z}^g$. The encoders are trained using a contrastive loss that maximizes the similarity between the embeddings $\vect{z}^t$ and $\vect{z}^g$ of paired transcripts and gestures.
        }
    \Description{}
    \label{fig:gesture-transcript_embedding_learning}
\end{figure*}

Our system takes the audio and transcript of a speech as input, synthesizing realistic, stylized full-body gestures that align with the speech content rhythmically and semantically. It allows using a short piece of text, namely a \emph{text prompt}, a video clip, namely a \emph{video prompt}, or a motion sequence, namely a \emph{motion prompt}, to describe a desired style. The gestures are then generated to embody the style as much as possible.

We build the system based on latent diffusion models \cite{rombach2022latentdiffusion}, which apply diffusion and denoising steps in a pretrained latent space. We learn this latent motion space using VQ-VAE \cite{van2017vqvae}, providing compact motion embeddings that ensure motion quality and diversity. As illustrated in \fig\ref{fig:system_overview}, our system is composed of two major components: (a) an end-to-end neural generator that accepts speech audio and text transcript as input and generates speech-matched gesture sequences using latent diffusion models; and (b) a CLIP-based encoder that extracts style embeddings from style prompts and integrates them into the diffusion models via an adaptive instance normalization (AdaIN) layer \cite{huang2017adain} to guide the style of the generated gestures. Furthermore, we learn a joint embedding space between corresponding gestures and transcripts using contrastive learning, which provides useful semantic cues for the generator and a semantic loss that effectively directs the generator to learn semantically meaningful gestures during training.

The system employs classifier-free diffusion guidance \cite{ho2022classifierfree} alongside with a self-supervised learning scheme,  enabling training on motion data without style labels. In the following sections, we will elaborate on the components and their training process within our system.
\section{Motion Representation}
\label{subsec:motion_representation}
A gesture motion $\vect{M}=[\vect{m}_k]_{k=1}^{K}$ is a sequence of poses, where $K$ denotes the motion length. Each pose $\vect{m}_k\in\mathbb{R}^{3+6J}$ consists of the displacement of the character and the rotations of its $J$ joints. We parameterize the rotations as 6D vectors \cite{zhou20196dvector}, though alternative rotation representations can potentially be used instead. The raw motion representation, however, often contains redundant information. To ensure motion quality and diversity while enabling fast inference, we follow recent successful systems \cite{ao2022rhythmicgesticulator,rombach2022latentdiffusion,dhariwal2020jukebox} and learn a compact motion representation using VQ-VAE \cite{van2017vqvae}.

Specifically, we train VQ-VAE as an encoder-decoder pair
\begin{align}
    \vect{Z} = \mathcal{E}_{\eqword{VQ}}(\vect{M})  \quad \Leftrightarrow \quad \vect{M} = \mathcal{D}_{\eqword{VQ}}(\vect{Z}) .
\end{align}
The encoder $\mathcal{E}_{\eqword{VQ}}$ converts $\vect{M}$ into a downsampled sequence of latent codes $\vect{Z}=[\vect{z}_l]_{l=1}^{L}$, where $\vect{z}_l\in\mathbb{R}^{C}$ and $C$ is the dimension of the latent space. We refer to the ratio $d=K/L$ as the encoder's downsampling rate, which is determined by the network structure. The decoder $\mathcal{D}_{\eqword{VQ}}$ operates on a quantized version of the latent space. It maintains a \emph{codebook} consisting of $N_{\eqword{VQ}}$ latent vectors. When reconstructing the original motion $\vect{M}$ from $\vect{Z}$, the decoder maps each $\vect{z}_k$ to its nearest codebook vector $\hatvect{z}_l$ and decodes the quantized latent sequence $\hatvect{Z}=[\hatvect{z}_l]_{l=1}^{L}$ into $\vect{M}$.

Our VQ-VAE model has a network structure similar to that of Jukebox \cite{dhariwal2020jukebox}, which consists of a cascade of 1D convolutional networks. The encoder $\mathcal{E}_{\eqword{VQ}}$ and decoder $\mathcal{D}_{\eqword{VQ}}$ are learned following the standard VQ-VAE training process \cite{van2017vqvae,dhariwal2020jukebox}. They are then frozen in the rest of training. Both the latent sequence $\vect{Z}$ and its quantized version $\hatvect{Z}$ are used as the motion representation by the other components of the system. Specifically, we learn the gesture-transcript joint embeddings on the quantized latent sequence $\hatvect{Z}$ in Section~\ref{sec:gesture-transcript_embedding_learning}, while the latent diffusion model synthesizes gesture motions as $\vect{Z}$ in Section~\ref{sec:stylized_co-speech_gesture_diffusion_model}.
\section{Gesture-Transcript Joint Embeddings}
\label{sec:gesture-transcript_embedding_learning}

The many-to-many mapping between speech content and gestures poses challenges in generating semantically correct motions. To alleviate this problem, we learn a joint embedding space for gestures and speech transcripts, enabling the discovery of semantic connections between the two modalities.

\subsection{Architecture}
As shown in \fig\ref{fig:gesture-transcript_embedding_learning}, we train two encoders, a gesture encoder $\mathcal{E}_G$ and a transcript encoder $\mathcal{E}_T$, to map the gesture motion and speech transcripts into the shared embedding space respectively. Both the encoders process the input speech in sentences. The speech transcripts are tokenized using the {T5 tokenizer \cite{xue2021mt5}} and temporally associated with the audio using the {Montreal Forced Aligner (MFA) \cite{mcauliffe2017mfa}}. This procedure also aligns the transcripts with the gestures. The speech data is subsequently segmented into sentences based on the transcripts. Following this, we compute:
\begin{align}
    \vect{Z}^t=\mathcal{E}_T(\vect{T}) , \quad  \quad \vect{Z}^g=\mathcal{E}_G(\hatvect{Z}) ,
\end{align}
where $\vect{T}\in \mathcal{W}^{L_t}$ denotes a tokenized transcript sentence parameterized as a sequence of word embeddings $w\in\mathcal{W}$, $\hatvect{Z}\in \mathbb{R}^{L_g \times C}$ is the quantized latent representation of the corresponding gesture sequence. The output of the encoders, $\vect{Z}^t\in\mathbb{R}^{L_t \times C_{{s}}}$ and $\vect{Z}^g\in\mathbb{R}^{L_g \times C_{{s}}}$, are sequences of feature vectors of the same dimension $C_{{s}}$. Note that the lengths of these sequences, $L_t$ and $L_g$, can be different.

In a speech, a semantic gesture and the utterance of its corresponding word or phrase often lack perfect alignment \cite{liang2022seeg}. This misalignment can confuse the encoder if the temporal correspondence between the two modalities is rigidly enforced during training. To alleviate this issue, we aggregate semantics-relevant information in each feature sequence via max pooling
\begin{align}
    \vect{z}^t = \eqword{max\_pooling}(\vect{Z}^t)  , \quad 
    \vect{z}^g = \eqword{max\_pooling}(\vect{Z}^g) .
\end{align}
Then $\vect{z}^t,\vect{z}^g \in \mathbb{R}^{C_{{s}}}$ are considered the embeddings of the transcripts and gestures, respectively. 

We employ a powerful pretrained language model, T5-base \cite{xue2021mt5}, as the text encoder $\mathcal{E}_T$. The motion encoder $\mathcal{E}_G$ is a 12-layer, 768-feature wide, encoder-only transformer with 12 attention heads, pretrained on the gesture dataset by predicting masked motions in a way similar to BERT \cite{devlin2019bert}. Both encoders are subsequently fine-tuned using contrastive learning, as detailed below.

\subsection{Contrastive Learning}
\begin{figure}[t]
    \centering
    \includegraphics[width=\linewidth]{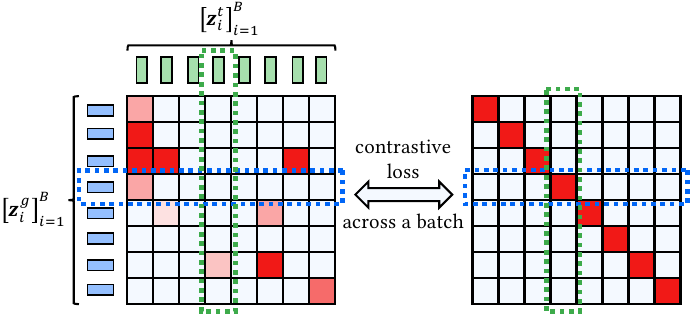}
    \caption{An illustration of the CLIP-style contrastive loss used to train the gesture and transcript encoders.}
    \Description{}
    \label{fig:contrastive_loss}
\end{figure}
We apply CLIP-style contrastive learning \cite{radford2021clip} to fine-tune the encoders. Given a batch of pairs of gesture and transcript embeddings $\mathcal{B}=\{(\vect{z}^t_i,\vect{z}^g_i)\}_{i=1}^{B}$, where $B$ is the batch size, the goal of the training is to maximize the similarity of the embeddings $(\vect{z}^t_i,\vect{z}^g_i)$ of the real pairs in the batch while minimizing the similarity of the incorrect pairs $(\vect{z}^t_i, \vect{z}^g_j)_{i\neq{}j}$. As illustrated in \fig\ref{fig:contrastive_loss}, this learning objective can be expressed as the sum of the gesture-to-text ($\text{g2t}$) cross entropy and the text-to-gesture ($\text{t2g}$) cross entropy computed across the batch. Formally, the loss function is
\begin{align}
    \mathcal{L}_{\eqword{contrast}} = \mathbb{E}_{\mathcal{B}\sim\mathcal{D}}&\Bigl[
        \mathcal{H}_{\mathcal{B}}\left(\vect{y}^{\eqword{g2t}}(\vect{z}^g_i), \vect{p}^{\eqword{g2t}}(\vect{z}^g_i)\right) 
        \nonumber \\ 
    &+ \mathcal{H}_{\mathcal{B}}\left(\vect{y}^{\eqword{t2g}}(\vect{z}^t_j), \vect{p}^{\eqword{t2g}}(\vect{z}^t_j)\right)\Bigr].
\end{align}
Each cross entropy $\mathcal{H}$ is computed between a one-hot encoding $\vect{y}$ and a softmax-normalized distribution $\vect{p}$. $\vect{y}$ specifies the true correspondence between the gestures and transcripts in the training batch $\mathcal{B}$. $\vect{p}$ computes the similarity between an embedding of one modality and those of the other modality. Specifically,
\begin{align}
    \label{eqn:multimodal_similarity}
    \vect{p}^{\eqword{g2t}}(\vect{z}^g_i) = \frac{\exp(\vect{z}^g_i \cdot \vect{z}^t_i / \tau)}{\sum^B_{j=1}\exp(\vect{z}^g_i \cdot \vect{z}^t_j / \tau)} 
    ,\quad
    \vect{p}^{\eqword{t2g}}(\vect{z}^t_j) = \frac{\exp(\vect{z}^t_j \cdot \vect{z}^g_j / \tau)}{\sum^B_{i=1}\exp(\vect{z}^t_j \cdot \vect{z}^g_i / \tau)},
\end{align}
where $\tau$ is the temperature of softmax.

In real data, there is often no corresponding gesture for an utterance, and a semantic feature may correspond to several different gestures. Such noisy correspondence could cause instability in contrastive learning. We employ the \emph{momentum distillation} (MoD) \cite{li2021albef} technique to alleviate this problem. The key idea of MoD is to learn from the pseudo-targets generated by a momentum model. During training, we maintain a momentum version of the encoders by updating their network parameters in an exponential-moving-average (EMA) manner. Then, we use the momentum models to calculate multimodal features $\tildevect{z}^t$ and $\tildevect{z}^g$ for the training gesture-transcript pairs and compute the pseudo-targets $\tildevect{p}^{\eqword{g2t}}$ and $\tildevect{p}^{\eqword{t2g}}$ by substituting these features into \eqn\eqref{eqn:multimodal_similarity}. The contrastive loss is then modified as
\begin{align}
    \mathcal{L}_{\eqword{contrast}}^{\eqword{MoD}} &= (1 - w_{\eqword{contrast}})\mathcal{L}_{\eqword{contrast}} \nonumber \\
    &+ w_{\eqword{contrast}}\mathbb{E}_{\mathcal{B} \sim \mathcal{D}}\Bigl[D_{KL}\left(\tildevect{p}^{\eqword{g2t}}(\tildevect{z}^g_i) || \vect{p}^{\eqword{g2t}}(\vect{z}^g_i)\right) \nonumber \\
    &+ D_{KL}\left(\tildevect{p}^{\eqword{t2g}}(\tildevect{z}^t_j) || \vect{p}^{\eqword{t2g}}(\vect{z}^t_j)\right)\Bigr],
\end{align}
where $D_{KL}(\cdot || \cdot)$ is the KL divergence and $w_{\eqword{contrast}}$ is set to $0.4$.

\subsection{Applications of the Joint Embeddings}
\begin{figure}[t]
    \centering
    \begin{subfigure}[t]{\linewidth}
        \centering
        \includegraphics[width=\linewidth]{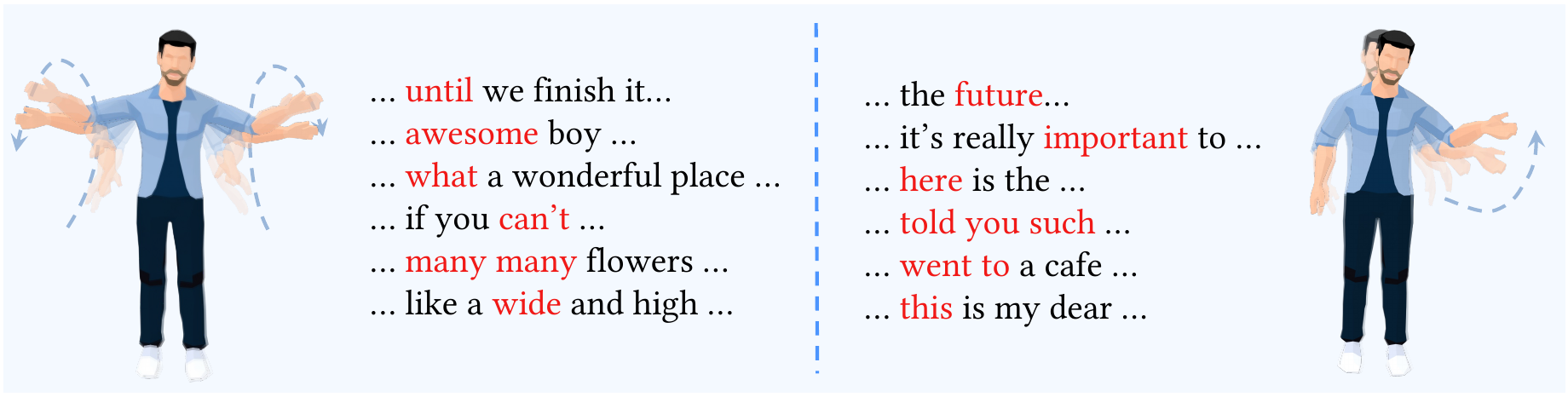}
        \caption{Transcripts retrieved based on example gestures. Note that a gesture can natually accompany several semantics.}
        \label{fig:motion-based_transcripts_retrieval}
    \end{subfigure} \\
    \vspace{5pt}
    \begin{subfigure}[t]{\linewidth}
        \centering
        \includegraphics[width=\linewidth]{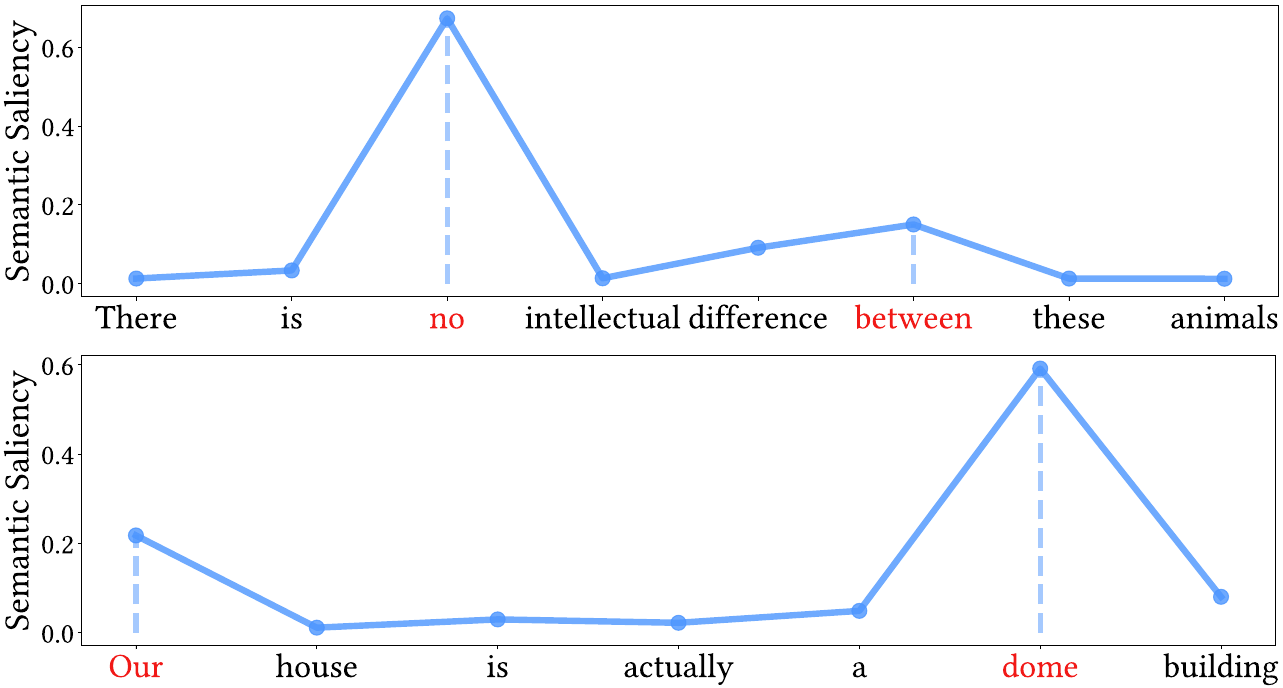}
        \caption{Semantic saliency curves of two sentences. The peaks of the curves indicate the words with high semantic importance which are likely to be accompanied by semantic gestures.}
        \label{fig:semantic_saliency}
    \end{subfigure}
    \caption{Applications of the gesture-transcript joint embeddings. (a) Motion-based transcripts retrieval . (b) Semantic saliency identification.}
    \label{fig:applications_of_gesture-transcript_embedding}
    \Description{}
\end{figure}
\begin{figure*}[t]
    \centering
    \includegraphics[width=\textwidth]{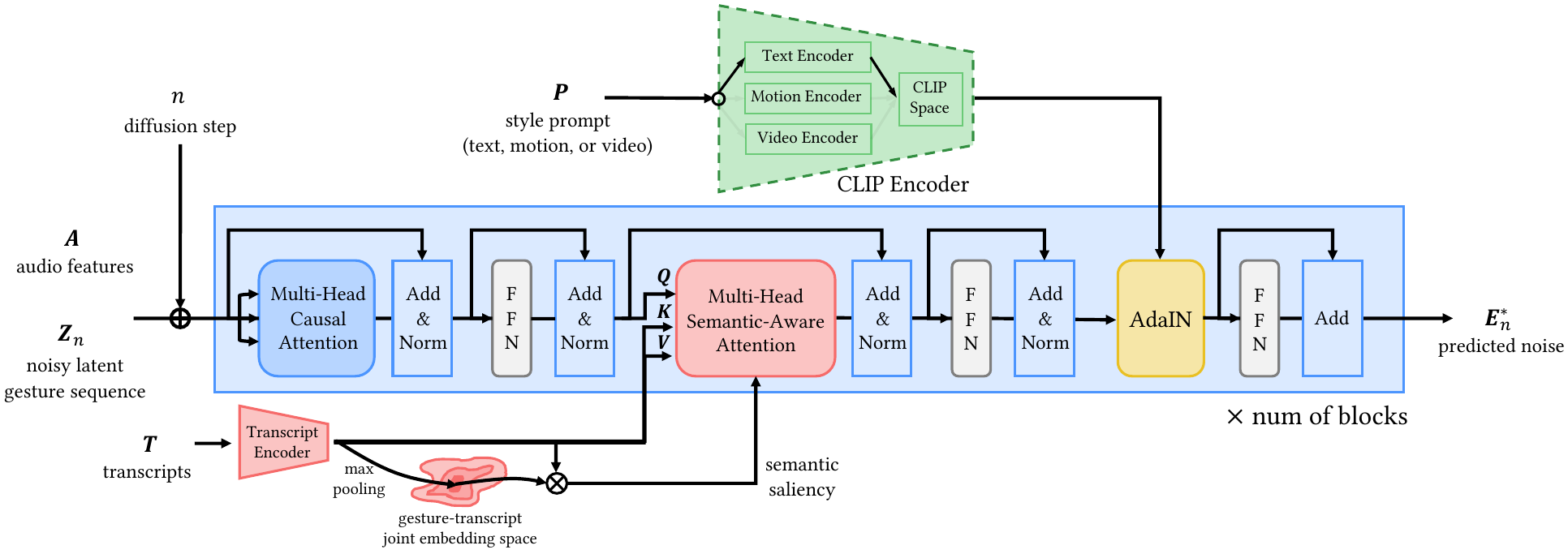}
    \caption{Architecture of the denoising network. The model is a multi-layer transformer with a causal attention structure. It takes the audio and transcript of a speech, along with a style prompt, as input and estimates the diffusion noise. Three CLIP-based encoders are learned to support different types of style prompts. The multimodal features are integrated  into the network at various stages through semantics-aware layers and AdaIN layers, respectively.
    \emph{Norm} refers to the layer normalization and \emph{FFN} is the feed-forward network.}
    \Description{}
    \label{fig:denoising_network}
\end{figure*}
The joint embedding space, along with the encoders, provides an efficient method for measuring semantic similarity between gestures and transcripts. To demonstrate its effectiveness, we map a gesture motion into this space and retrieve the closest sentences from the transcript dataset based on the cosine distance of the embeddings. \fig\ref{fig:motion-based_transcripts_retrieval} shows some results. It can be observed that the retrieved sentences may have various meanings, but can all be naturally paired with the query gestures.

Besides, the computation of the embeddings involves the max pooling operator, which aggregates the most semantics-relevant information. Consequently, we can estimate the saliency of each pose or word in a gesture sequence or a transcript sentence, respectively, using the embeddings. Specifically, given a sentence $\vect{T}$ along with its encoded feature sequence $\vect{Z}^t$ and embedding vector $\vect{z}^t$, we compute the semantic saliency of each word as
\begin{align}
    \label{eqn:semantic-saliency}
    \vect{s}^t = \eqword{softmax}\left(\vect{Z}^t \cdot \vect{z}^t\right).
\end{align}
As illustrated in \fig\ref{fig:semantic_saliency}, the words with high semantic importance that are likely to be accompanied by semantic gestures will exhibit high saliency scores. This information can be considered as an important semantic cue, which will be used in our system to guide the gesture generator  in creating semantically correct gestures.
\section{Stylized Co-Speech Gesture Diffusion Model}
\label{sec:stylized_co-speech_gesture_diffusion_model}
The core of our system is a conditional latent generative model $\mathcal{G}$ which synthesizes a sequence of latent gesture codes $\vect{Z}=[\vect{z}_l]_{l=1}^{L}$ conditioned on a speech and a style prompt. The latent sequence $\vect{Z}$ is then decoded into gestures using the VQ-VAE decoder $\mathcal{D}_{\eqword{VQ}}$ learned in Section~\ref{subsec:motion_representation}. Formally, the generator $\mathcal{G}$ computes
\begin{align}
    \vect{Z} = \mathcal{G}(\vect{A}, \vect{T}, \vect{P})
\end{align}
where $\vect{A}$ and $\vect{T}$ denote the audio and transcript of the speech respectively and $\vect{P}$ is the style prompt. The speech audio $\vect{A}=[\vect{a}_i]_{i=1}^{L}$ is parameterized as a sequence of acoustic features, sampled to match the length of the gesture representation. Each $\vect{a}_i$ encodes the onsets and amplitude envelopes that reflect the beat and volume of speech, respectively. The speech transcript $\vect{T}$ is preprocessed as described in Section~\ref{sec:gesture-transcript_embedding_learning}. The generator $\mathcal{G}$ uses $\vect{A}$ to infer low-level gesture styles such as rhythm and stress, $\vect{T}$ for the semantic-level features, and $\vect{P}$ to determine the overall style of the gestures.

During inference, the generator $\mathcal{G}$ is reformulated as an autoregressive model, where a gesture is determined by not only the speech context and style prompt but also the previous motion. Formally, the latent sequence $\vect{Z}=[\vect{z}_l]_{l=1}^{L}$ is generated as
\begin{align}
    \label{eqn:autoregressive_generative_model}
    {\vect{z}_l^*} = \mathcal{G}([\vect{z}_i^*]_{i=1}^{l-1}, [\vect{a}_i]_{i=1}^{l+\delta^a}, \vect{T}, \vect{P}),
\end{align}
where we use the asterisk ($*$) to indicate quantities already generated by $\mathcal{G}$. Note that the generator leverages $\delta^a$ frames of future audio features to determine the current gestures.

\subsection{Latent Diffusion Models}
The generator $\mathcal{G}$ is based on the latent diffusion model \cite{rombach2022latentdiffusion}, which is a variant of diffusion models that applies the forward and reverse diffusion processes in a pretrained latent feature space. The \emph{diffusion process} is modeled as a Markov noising process. Starting from a latent gesture sequence $\vect{Z}_0$ drawn from the gesture dataset, the diffusion process progressively adds Gaussian noise to the real data until its distribution approximates $\mathcal{N}(\vect{0}, \vect{I})$. The distribution of the latent sequences thus evolves as
\begin{align}
    q(\vect{Z}_n | \vect{Z}_{n-1}) = \mathcal{N}(\sqrt{\alpha_n}\vect{Z}_{n-1}, (1-\alpha_n)\vect{I}),
\end{align}
where $\vect{Z}_n$ is the latent sequence sampled at diffusion step $n$, $n\in\{1, \dots, N\}$, and $\alpha_n$ is determined by the variance schedules. In contrast, the \emph{reverse diffusion process}, or the \emph{denoising process}, estimates the added noise in a noisy latent sequence. Starting from a sequence of random latent codes $\vect{Z}_N \sim \mathcal{N}(\vect{0}, \vect{I})$, the denoising process progressively removes the noise and recovers the original motion $\vect{Z}_0$.

To achieve conditional gesture generation, we train a network $\vect{E}_{\theta}$, the so-called \emph{denoising network}, to predict the noise based on the noisy motion codes, the diffusion step, the speech context, and the style prompt. This network can be formulated as
\begin{align}
    \label{eqn:denoising_network}
    \vect{E}^{*}_n = \vect{E}_{\theta}(\vect{Z}_n, n, \vect{A}, \vect{T}, \vect{P}).
\end{align}
During inference, the generator $\mathcal{G}$ leverages the sampling algorithm of DDPM~\cite{ho2020ddpm} to synthesize gestures. It first draws a sequence of random latent codes $\vect{Z}_N^*\sim{}\mathcal{N}(\vect{0}, \vect{I})$ then computes a series of denoised sequences $\{\vect{Z}_n^*\},{n=N-1,\dots,0}$ by iteratively removing the estimated noise $\vect{E}_n^*$ from $\vect{Z}_n^*$. The entire process is carried out in an autoregressive manner. Specifically, we first sample the initial latent code $\textbf{z}_{1(N)}$ for the first frame and denoise it for $N$ steps, obtaining the generated gesture code $\textbf{z}_{1(0)}^{*}$. Next, we generate the second frame gesture code $\textbf{z}_{2(0)}^{*}$ by denoising an initial code $\textbf{z}_{2(N)}$ based on the previous code $\textbf{z}_{1(0)}^{*}$ and other conditions. This process is repeated autoregressively to generate a gesture sequence, with the previous codes $[\textbf{z}_{i(n)}]^{l-1}_{i=1}$ being replaced by the generated codes $[\textbf{z}_{i(0)}^{*}]^{l-1}_{i=1}$ for each frame $l$. This strategy is naturally extended to generate long sequences conditioned on previously generated gestures. Finally, the latent codes $\vect{Z}_0^*$ are decoded into the gesture motion. 

As illustrated in \fig\ref{fig:denoising_network}, our denoising network has a transformer architecture. We employ the causal attention layer proposed by \citet{vaswani2017transformer} that only allows the intercommunication of the current and preceding data for the causality. This architecture can be easily transformed into the autoregressive model in \eqn\eqref{eqn:autoregressive_generative_model}. Note that we extend the definition of \emph{current data} when dealing with audio features by including $\delta^a$ future frames.

The denoising network fuses the multimodal conditions $(\vect{A}, \vect{T}, \vect{P})$ in a hierarchical manner: first, low-level audio features that relate to the speech rhythm and stress are integrated by concatenating $\vect{A}$ with the noisy latent sequence $\vect{Z}_n$; then, high-level transcript features $\vect{T}$ that correspond to the speech semantics are incorporated via a \emph{semantics-aware attention layer}; lastly, the style prompt $\vect{P}$ is included through a \emph{CLIP-guided AdaIn layer} to control the overall style of the generated gestures.

\subsubsection{Semantics-Aware Attention Layer}
Inspired by recent successful attention-based multimodal systems \cite{jaegle2021perceiver,rombach2022latentdiffusion}, we develop a semantics-aware attention layer based on the cross-attention mechanism \cite{vaswani2017transformer} to incorporate the input transcript $\vect{T}$. Specifically, we first extract the transcript features $\vect{Z}^t$ from $\vect{T}$ using the pretrained text encoder $\mathcal{E}_T$ as described in Section~\ref{sec:gesture-transcript_embedding_learning} and compute the semantic saliency $\vect{s}^t$ using \eqn\eqref{eqn:semantic-saliency}. Then, we project $\vect{Z}^t$ to the \emph{key} $\vect{K} \in \mathbb{R}^{L_t \times C_t}$ and \emph{value} $\vect{V} \in \mathbb{R}^{L_t \times C_t}$ of the attention mechanism using learnable projection matrices and calculate the \emph{query} $\vect{Q} \in \mathbb{R}^{L \times C_t}$ using the intermediate features of the denoising network. Finally, the semantics-aware attention layer is implemented as
\begin{align}
    \eqword{Attention}(\vect{Q}, \vect{K}, \vect{V}) = \eqword{softmax}(\frac{\vect{Q}\vect{K}^T}{\sqrt{C_t}} \cdot \vect{S}^t) \cdot \vect{V},
\end{align}
where $\vect{S}^t \in \mathbb{R}^{L \times L_t}$ is the temporally broadcasted semantic saliency matrix of $\vect{s}^t$ that guides the network to pay extra attention to the semantically important words.

\subsubsection{CLIP-Guided AdaIN Layer}
\label{subsec:CLIP-guided-adain}
We employ an adaptive instance normalization (AdaIN) layer \cite{huang2017adain} to inject the information of the style prompts into the denoising network. Specifically, we leverage a pretrained CLIP encoder $\mathcal{E}_{\eqword{CLIP}}$ to convert the input style prompt into a style embedding $\vect{z}^s \in \mathbb{R}^{C_{\eqword{CLIP}}}$. Then, we learn a MLP network to map the style embedding $\vect{z}^s$ to $2 C_{\eqword{ada}}$ parameters that modify the per-channel mean and variance of the AdaIn layer with $C_{\eqword{ada}}$ channels.

We employ the text encoder $\mathcal{E}_{\eqword{CLIP-T}}$ of the CLIP model \cite{radford2021clip} for the text prompts and the motion encoder $\mathcal{E}_{\eqword{CLIP-M}}$ of the MotionCLIP model \cite{tevet2022motionclip} for the motion prompts. We further develop a CLIP-based video encoder $\mathcal{E}_{\eqword{CLIP-V}}$ for the video prompts, which consists of a pretrained CLIP image encoder \cite{radford2021clip} followed by a $6$-layer transformer that temporally aggregates the image features of the video into an embedding $\vect{z}^s$ in the CLIP space. Note that all these CLIP encoders are pretrained in a separate stage and their weights are frozen when training the denoising network.

\subsection{Training}
\label{subsubsec:training_of_denosing_network}
Following the standard training process of denoising diffusion models \cite{ho2020ddpm,rombach2022latentdiffusion}, we train the denoising network $\vect{E}_{\theta}$ by drawing random tuples $(\vect{Z}_0,n,\vect{A},\vect{T},\vect{P})$ from the training dataset, then corrupting $\vect{Z}_0$ into $\vect{Z}_n$ by adding random Gaussian noises $\vect{E}$, applying denoising steps to $\vect{Z}_n$, and optimizing the loss 
\begin{align}
    \mathcal{L}_{\eqword{net}} = 
    w_{\eqword{noise}}\mathcal{L}_{\eqword{noise}} +
    w_{\eqword{semantic}}\mathcal{L}_{\eqword{semantic}} +
    w_{\eqword{style}}\mathcal{L}_{\eqword{style}}.
\end{align}
Specifically, the ground-truth gesture motion $\vect{M}_0$, along with its latent representation $\vect{Z}_0$, and the speech audio $\vect{A}$ and transcript $\vect{T}$ are extracted from the same speech sentence. $n$ is drawn from the uniform distribution $\mathcal{U}\{1,N\}$. We do not assume that the gesture dataset contains detailed style labels. Instead, we consider the motion clip $\vect{M}_P$ of a random length but encompassing $\vect{M}_0$ as the style prompt $\vect{P}$.

We first calculate the standard noise estimation loss of the diffusion models \cite{ho2020ddpm} defined as
\begin{align}
    \mathcal{L}_{\eqword{noise}} = \norm{\vect{E} - \vect{E}_{\theta}(\vect{Z}_n, n, \vect{A}, \vect{T}, \vect{P})}_{2}^{2},
\end{align}
Then, we leverage a semantic loss to ensure the semantic correctness of the generated gestures. This loss is defined in the gesture-transcript joint embedding space learned in Section \ref{sec:gesture-transcript_embedding_learning}. Specifically,
\begin{align}
    \mathcal{L}_{\eqword{semantic}} = 1 - \cos(\vect{z}^g_0, \vect{z}^{g*}_0),
    \label{eqn:sem_loss}
\end{align}
where $\cos(\cdot, \cdot)$ is the cosine distance, $\vect{z}^g_0$ and $\vect{z}^{g*}_0$ are the gesture encodings of the ground-truth and generated motions, respectively, computed using the gesture encoder $\mathcal{E}_G$ pretrained in Section \ref{sec:gesture-transcript_embedding_learning}.
At last, we employ a perceptual loss to encourage the generator to follow the style prompts. This style loss is defined as
\begin{align}
    \mathcal{L}_{\eqword{style}} = 1 - \cos(\mathcal{E}_{\eqword{CLIP-M}}(\vect{M}_0), \mathcal{E}_{\eqword{CLIP-M}}(\vect{M}^{*}_0)),
    \label{eqn:style_loss}
\end{align}
where $\mathcal{E}_{\eqword{CLIP-M}}$ is the pretrained motion encoder, and $\vect{M}^{*}_0=\mathcal{D}_{\eqword{VQ}}(\vect{Z}_0^*)$ is the generated gestures. As suggested by \citet{kim2022CVPRDiffusionClip}, in each training iteration, we choose a random starting step $n$ and apply a complete denoising process to obtain $\textbf{Z}_{0}^{*}$, which is then used to compute the perceptual losses $\mathcal{L}_{\eqword{semantic}}$ and $\mathcal{L}_{\eqword{style}}$.

We utilize the classifier-free guidance \cite{ho2022classifierfree} to train our model. Specifically, we let $\vect{E}_{\theta}$ learn both the style-conditional and unconditional distributions by randomly setting $\vect{P} = \varnothing$ and thus disabling the AdaIN layer by $10\%$ chance during training. At inference time, the predicted noise is computed using 
\begin{align}
    \vect{E}^*_n&= s \vect{E}_{\theta}(\vect{Z}_n, n, \vect{A}, \vect{T}, \vect{P}) + (1-s)  \vect{E}_{\theta}(\vect{Z}_n, n, \vect{A}, \vect{T}, \varnothing)
\end{align}
instead of \eqn\eqref{eqn:denoising_network}. This scheme further allows us to control the effectiveness of the style prompt $\vect{P}$ by adjusting the scale factor~$s$.

\subsubsection{CLIP Video Encoder}
We develop a CLIP-based video encoder $\mathcal{E}_{\eqword{CLIP-V}}$ in Section~\ref{subsec:CLIP-guided-adain} to enable video clips as the style prompts. $\mathcal{E}_{\eqword{CLIP-V}}$ encapsulates a pretrained CLIP image encoder \cite{radford2021clip} whose weights are frozen and a learnable transformer network. To learn $\mathcal{E}_{\eqword{CLIP-V}}$, we render a random motion sequence  $\vect{M}$ into a video and optimize the loss function
\begin{align}
    \label{eqn:clip-video-encoder-loss}
    \mathcal{L}_{\eqword{video}} = 1 - \cos(\eqword{sg}(\mathcal{E}_{\eqword{CLIP-M}}(\vect{M})), \mathcal{E}_{\eqword{CLIP-V}}(\mathcal{R}(\vect{M}; \vect{r}))) ,
\end{align}
where $\eqword{sg}$ represents the \emph{stop gradient} operator that prevents the gradient from backpropagating through it, $\mathcal{R}$ denotes the rendering operator that renders $\vect{M}$ into a video of human skeleton poses, with camera parameters $\vect{r}$ configured similarly to \cite{aberman2020adain}, and $\mathcal{E}_{\eqword{CLIP-M}}$ is the pretrained motion encoder. This loss function ensures $\mathcal{E}_{\eqword{CLIP-V}}$ to map video clips into the same shared CLIP embedding space. Interestingly, although the video encoder is fine-tuned using only synthetic videos, we find that it can still extract meaningful semantic information from real-world videos in practice. We attribute this robustness to the pretrained image decoder, which was trained on a large dataset \cite{radford2021clip}.

\subsection{Style Control of Body Parts}
Inspired by \cite{zhang2022motiondiffuse}, we extend our system to allow fine-grained styles control on individual body parts using \emph{noise combination}. Considering a partition $\mathcal{O}$ of the character's body, where each body part $o\in\mathcal{O}$ consists of several joints, we learn $O=|\mathcal{O}|$ individual motion VQ-VAEs to represent the motions of each body part as latent codes $\vect{Z}^{o} = \mathcal{E}_{\eqword{VQ}}^o(\vect{M}^o)$. The full-body motion codes $\vect{Z}^{\mathcal{O}} \in \mathbb{R}^{O \times (L \times C)}$ is then computed by stacking the motion codes of each body part. We then train a new latent diffusion model $\vect{E}_{\theta}$ based on $\vect{Z}^{\mathcal{O}}$ in the same way as introduced in the previous sections. At inference time, we predict full-body noises $\{\vect{E}_{n,o}^*\}_{o\in\mathcal{O}}$ conditioned on a set of style prompts $\{\vect{P}_o\}_{o\in\mathcal{O}}$ for every body part, where each $\vect{E}_{n,o}^*= \vect{E}_{\theta}(\vect{Z}_n^{\mathcal{O}}, n, \vect{A}, \vect{T}, \vect{P}_o)$. These noises can be simply fused as $\vect{E}_n^* = \sum_{o\in\mathcal{O}}\vect{E}_{n,o}^* \cdot M_o$, where $\{M_o\}_{o\in\mathcal{O}}$ are binary masks indicating the partition of bodies in $\mathcal{O}$. To achieve better motion quality, we add a smoothness item to the denoising direction as suggested by~\citet{zhang2022motiondiffuse}, which is
\begin{align}
    \vect{E}_n^* = \sum_{o\in\mathcal{O}}\Bigl(\vect{E}_{n,o}^* \cdot M_o\Bigr) + 
    w_{\eqword{body}} \nabla_{\vect{Z}_n^{\mathcal{O}}}{\Bigl(\sum_{i,j \in\mathcal{O}, i\neq{}j }\norm{\vect{E}_{n,i}^* - \vect{E}_{n,j}^*}_2\Bigr)},
\end{align}
where $\nabla$ denotes the gradient operator. $w_{\eqword{body}}$ is set to 0.01.
\section{Evaluation}
\label{sec:results}
In this section, we first describe the setup of our system then evaluate our results, compare them with other systems, introduce several potential applications of our system, and validate various design choices of our framework through ablation study.

\subsection{System Setup}
\label{subsec:system_setup}

\subsubsection{Data}
We train and test our system on two high-quality speech-gesture datasets: \emph{ZeroEGGS} \cite{ghorbani2022zeroeggs} and \emph{BEAT} \cite{liu2021beatdataset}. The ZeroEGGS dataset contains two hours of full-body motion capture and audio from monologues performed by an English-speaking female actor in $19$ different styles. We acquire the synchronized transcripts using an automatic speech recognition (ASR) tool \cite{alibaba2009asr}. The BEAT dataset contains $76$ hours of multimodal speech data, including audio, transcripts, and full-body motion captured from $30$ speakers performing in eight emotional styles and four different languages. We only use the speech data of English speakers, which amounts to about $35$ hours in total.

\subsubsection{Settings}
Our system generates motions at $60$ frames per second. We train the motion VQ-VAE (Section \ref{subsec:motion_representation}) with a downsampling rate $d = 8$, a batch size of $32$, and $C = 512$. To learn the gesture-transcript embedding space (Section \ref{sec:gesture-transcript_embedding_learning}), the values of $C_s$, $\tau$, and $B$ are set to $768$, $0.07$, and $32$, respectively. As for the diffusion module (Section \ref{sec:stylized_co-speech_gesture_diffusion_model}), the denoising network is based on a $12$-layer, $768$-feature wide, encoder-only transformer with $12$ attention heads. The number of diffusion steps is $N=1000$, the training batch size is $128$ per GPU, and the parameters $\delta^a$, $C_{\eqword{CLIP}}$, $C_{\eqword{ada}}$, $w_{\eqword{noise}}$, $w_{\eqword{semantic}}$, $w_{\eqword{style}}$ are set to $8$, $768$, $768$, $1.0$, $0.1$, and $0.07$, respectively. We train the motion VQ-VAE using regular speech clips of $4$ seconds in length and other models with sentence-level clips ranging from $1$~second to $15$~seconds in length. We train all these models using two NVIDIA Tesla V100 GPUs for about five days. During inference, it takes our system about 1.5 seconds to generate an 1-second (60 frames) gesture clip on a single Tesla V100 GPU.

\subsection{Results}
\begin{figure*}[t]
    \centering
    \includegraphics[width=\textwidth]{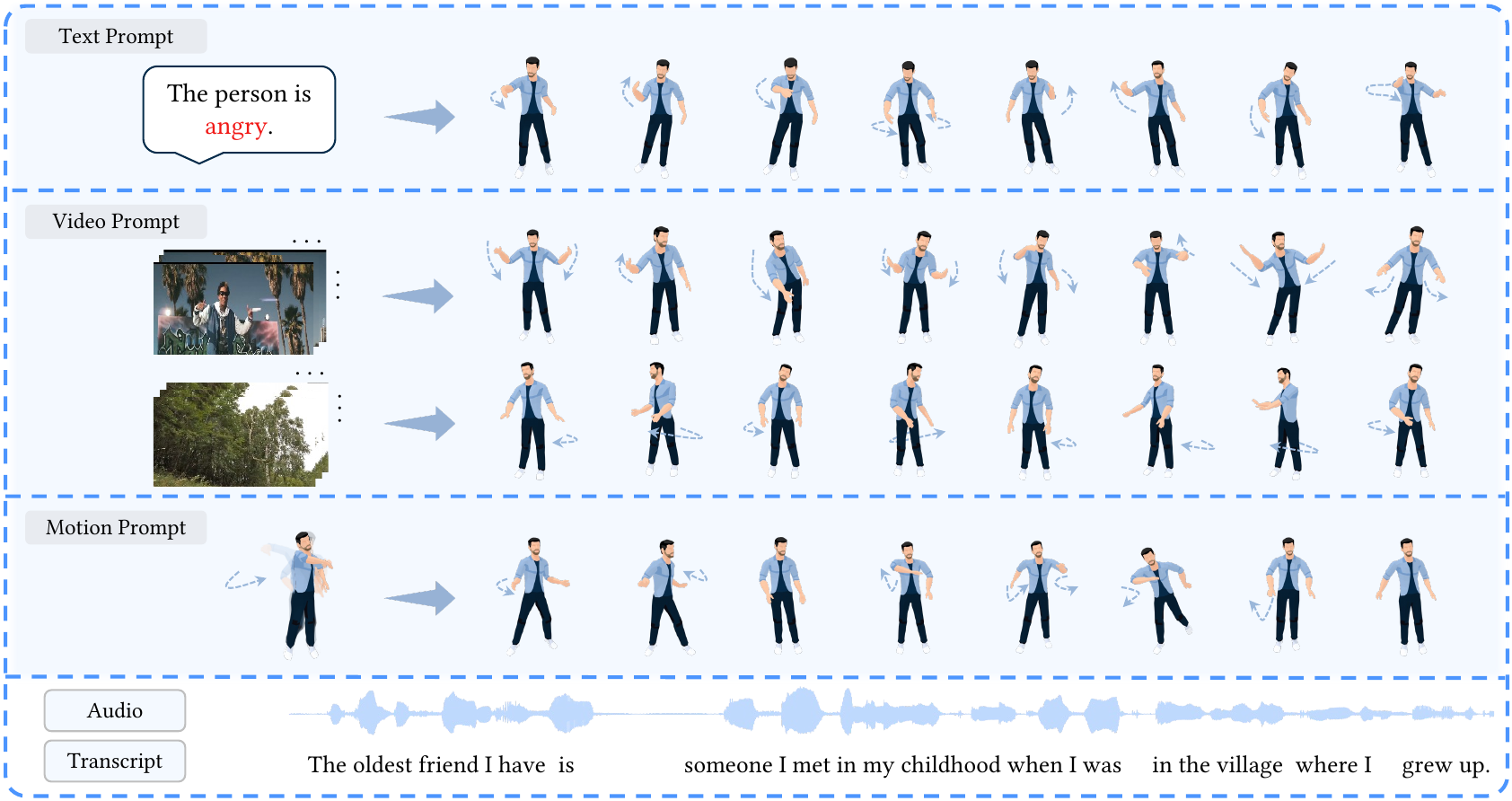}
    \caption{Gestures synthesized by our system conditioned on three different types of style prompts (text, video, and motion) for the same speech. The character performs \emph{angry} gestures when given the text prompt \emph{the person is angry}. The \emph{Hip-hop} style gestures are imitated from a Hip-hop music video \cite{khalifa2011hippopvideo}. Semantic information of a non-human video such as \emph{trees sway with the wind} \cite{rijavec2018windtreevideo} can also be perceived by our system and guides the character to sway from side to side. As for the motion prompt, our system successfully generates \emph{arm-up} gestures similar to the motion example.}
    \Description{}
    \label{fig:results_ours}
\end{figure*}
\begin{figure*}[t]
    \centering
    \includegraphics[width=\linewidth]{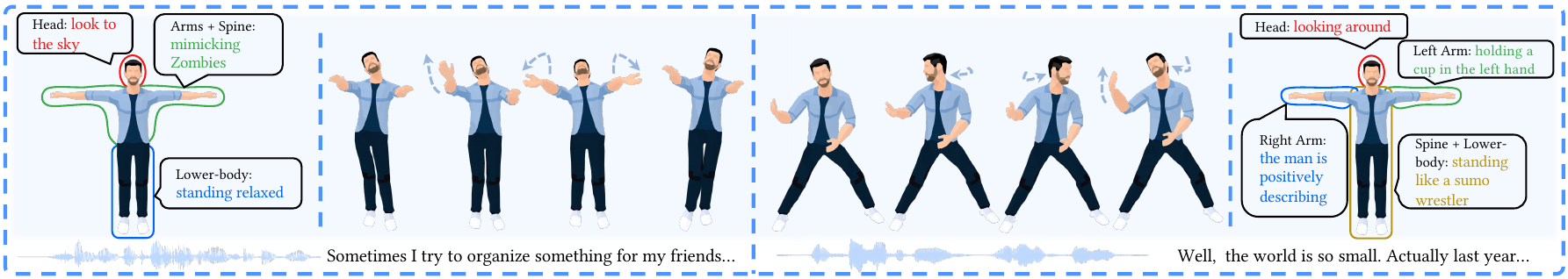}
    \caption{Results of body part-level style control. A set of style prompts are applied to different body parts to achieve fine-grained style control.}
    \Description{}
    \label{fig:sub-body_style_control}
\end{figure*}
\begin{figure}[t]
    \centering
    \includegraphics[width=\linewidth]{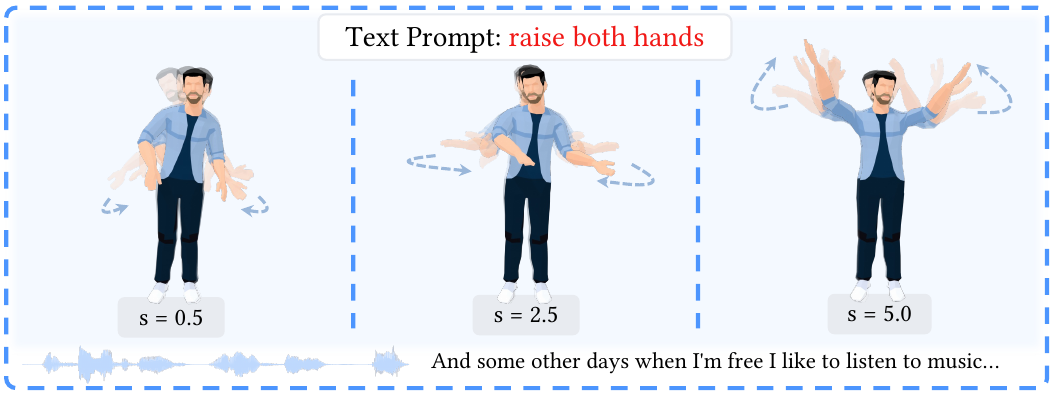}
    \caption{Effect of the input style prompt controlled by adjusting the scale $s$ of the classifier-free guidance. A larger $s$ results in a more pronounced style effect.}
    \Description{}
    \label{fig:effect_of_style_prompt}
\end{figure}
\begin{figure}[t]
    \centering
    \includegraphics[width=\linewidth]{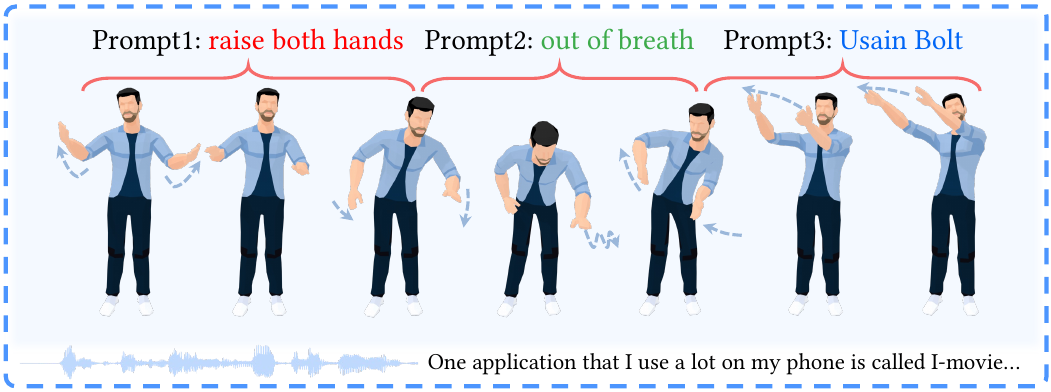}
    \caption{Results of the time-varied style control by changing style prompt at each sentence.}
    \Description{}
    \label{fig:time-varied_style_control}
\end{figure}
\fig\ref{fig:results_ours} shows the visualization results of our system generating gestures conditioned on three different types of style prompts (text, video, and motion), with the test speech taken from the ZeroEGGS dataset. Our system successfully generates realistic gestures with reasonable styles, as required by the corresponding prompts. The character performs \emph{angry} gestures when given the text prompt \emph{the person is angry}. The \emph{Hip-hop} style gestures are imitated from a Hip-hop music video \cite{khalifa2011hippopvideo}. Semantic information from a non-human video, such as \emph{trees sway with the wind} \cite{rijavec2018windtreevideo}, can also be perceived by our system, guiding the character to sway from side to side. As for the motion prompt, our system successfully generates \emph{arm-up} gestures similar to the motion example.

\fig\ref{fig:sub-body_style_control} demonstrates the results of body part-aware style control using our system. We employ different prompts to control the styles of various body parts. The resulting motions produce these styles while maintaining a natural coordination among the body parts.

As discussed in Section \ref{subsubsec:training_of_denosing_network}, the scale factor $s$ of the classifier-free guidance scheme controls the effect of the input style prompt. Increasing $s$ will enhance or even exaggerate the given style, which can be seen in \fig\ref{fig:effect_of_style_prompt}, where the hands of the character rise higher when increasing~$s$ with the text prompt \emph{raise both hands}.

Our system allows the style of gestures to change with every sentence. As shown in \fig\ref{fig:time-varied_style_control}, we change the style prompts from \emph{raise both hands} to \emph{out of breath}, and then to \emph{Usain Bolt}. The generated gestures accurately match the styles and maintain smooth transitions between different styles.

Moreover, \fig\ref{fig:application_style_prompt_generation} shows the visualization results of comparing style-conditional synthesis (the first row) with style-unconditional synthesis (the second row). Specifying style prompts for each sentence can effectively guide the character's performance, making the co-speech gestures more vivid.
\begin{table*}[t]
    \centering
    \caption{Average scores of user study with $95\%$ confidence intervals. \emph{Our system without transcript input (w/o transcript)} excludes the $\mathcal{L}_{\eqword{semantic}}$ while the semantic-aware attention layer is replaced by the causal self-attention layer. Then the generator is retrained to synthesize gestures based solely on the audio. 
    For \emph{style correctness (w/ dataset label)}, the text prompt contains the style label of the dataset and only describes the motion style that appears in the dataset. Note that we do not use any style label for training. Meanwhile, the text prompt utilized in the \emph{style correctness (w/ random prompt)} test does not contain any style label of the dataset, and the motion style that the text prompt describes may not exist in the dataset.}
    \label{tab:user_study}

    \newcolumntype{Y}{>{\raggedleft\arraybackslash}X}
    \newcolumntype{Z}{>{\centering\arraybackslash}X}
    \begin{tabularx}{\linewidth}{llYYYY}
        \toprule
        Dataset & System & Human Likeness $\uparrow$ & Appropriateness $\uparrow$ & \makecell{Style Correctness $\uparrow$ \\ (w/ dataset label)} & \makecell{Style Correctness $\uparrow$ \\ (w/ random prompt)} \\ 
        \toprule
        \multirow{4}*{BEAT} & GT & $0.47 \pm 0.08$ & $0.73 \pm 0.08$ & - & - \\
        \cline{2-6}
        & CaMN & $-0.99 \pm 0.12$ & $-1.06 \pm 0.10$ & - & - \\
        & Ours (w/o transcript) & $0.23 \pm 0.10$ & $-0.33 \pm 0.07$ & - & - \\
        & Ours & $\bm{0.31 \pm 0.07}$ & $\bm{0.51 \pm 0.07}$ & - & - \\
        
        \midrule
        \multirow{3}*{ZeroEGGS} & ZE & $-0.25 \pm 0.10$ & $-1.33 \pm 0.12$ & - & - \\
        & MD-ZE & - & - & $-1.65 \pm 0.15$ & $-1.62 \pm 0.17$ \\
        & Ours & $\bm{0.25 \pm 0.10}$ & $\bm{1.33 \pm 0.12}$ & $\bm{1.65 \pm 0.15}$ & $\bm{1.62 \pm 0.17}$ \\
        \bottomrule
    \end{tabularx}

\end{table*}

\subsection{Comparison}
Following the convention of recent gesture generation research \cite{alexanderson2020stylegesture,ghorbani2022zeroeggs,ao2022rhythmicgesticulator}, we evaluate the generated motions through a series of user studies. Quantitative evaluations are also conducted, with results provided in later sections as a reference.

\subsubsection{Baselines}
The BEAT dataset is released with a baseline approach, Cascaded Motion Network (CaMN), which takes the audio, transcript, emotion labels, speaker ID, and facial blendshape weights as inputs to generate gestures using a cascaded architecture. Here, we ignore the fingers and facial weights of the generated motion. Similarly, the ZeroEGGS dataset also provides a baseline, the ZeroEGGS algorithm (ZE), which encodes speech audio and a motion exemplar to synthesize target stylized gestures. ZE achieves the best performance among all deep-based models in the $2022$ GENEA Challenge \cite{yoon2022genea}. Because no model in previous studies supports text prompt-conditioned synthesis, we construct a baseline, MD-ZE, by combining a powerful text prompt-to-motion model, MotionDiffuse (MD) \cite{zhang2022motiondiffuse}, and ZeroEGGS. Specifically, MD first transfers the given text prompt into a motion exemplar, then ZE generates target stylized gestures based on the motion exemplar and input speech. For more implementation details of these baselines, please refer to Appendix \ref{sec:implementation_details_of_baselines}.
\begin{figure*}[t]
    \centering
    \includegraphics[width=\linewidth]{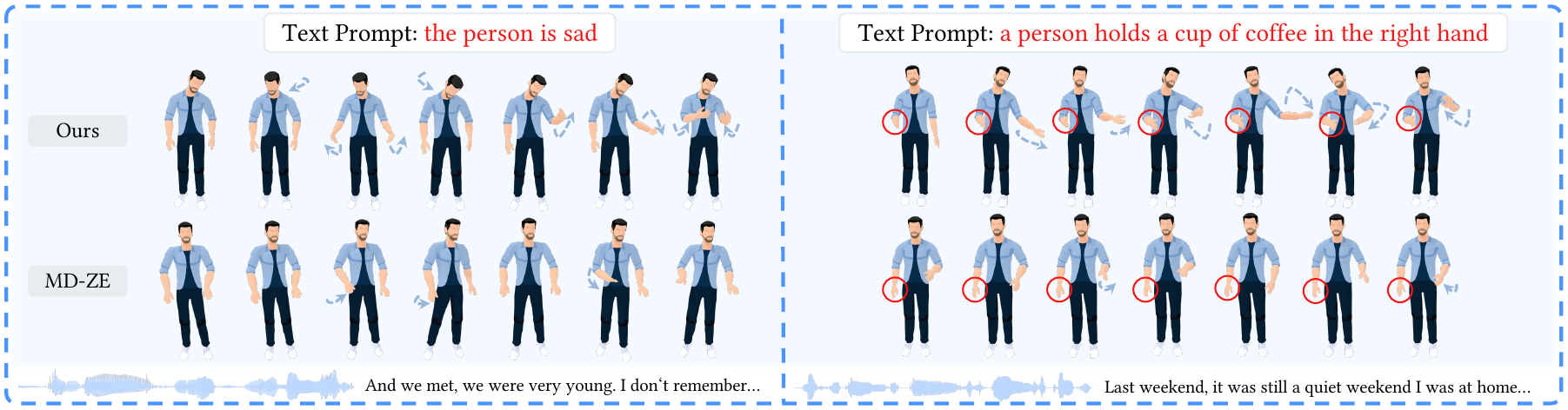}
    \caption{Qualitative comparison between our system and MD-ZE \cite{zhang2022motiondiffuse, ghorbani2022zeroeggs} using speech excerpts from the user study. The left text prompt displays the style label (angry) from the dataset, while the right prompt describes a motion style absent in the dataset.}
    \Description{}
    \label{fig:comparison_results}
\end{figure*}
\begin{figure*}[t]
    \centering
    \includegraphics[width=\textwidth]{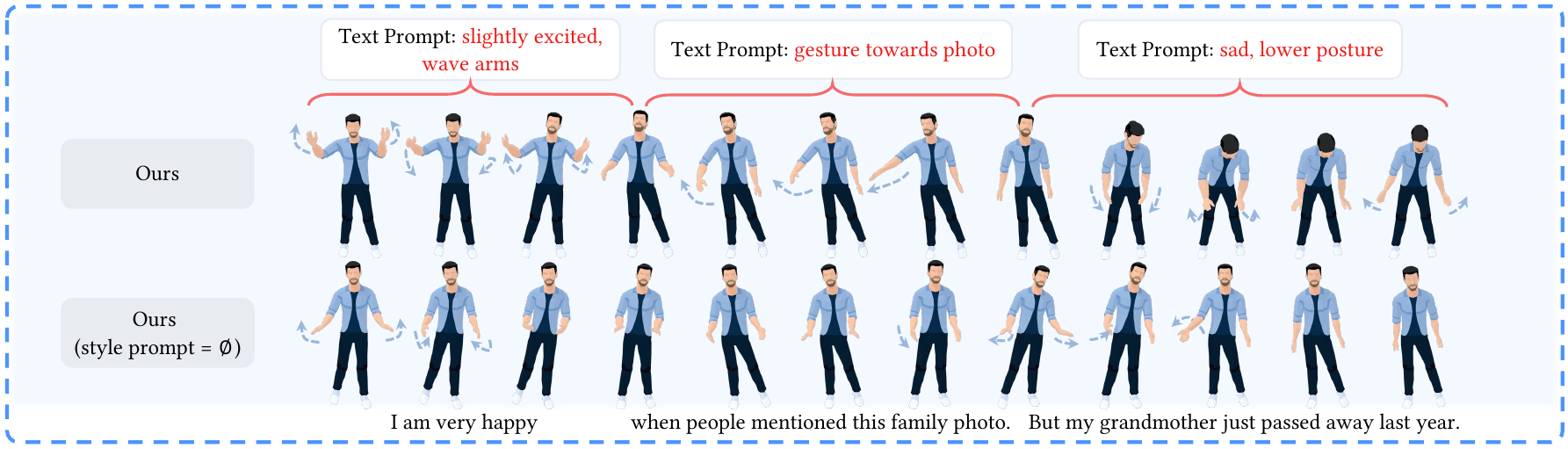}
    \caption{Qualitative comparison between style-conditional synthesis (first row) and style-unconditional synthesis (second row). Co-speech gestures are improved by providing a style prompt for each sentence within the speech transcript.
    }
    \Description{}
    \label{fig:application_style_prompt_generation}
\end{figure*}

\subsubsection{User Study}
\label{subsubsec:user_study}
We conduct user studies using pairwise comparisons, similar to the method described in \cite{alexanderson2022diffusiongesture}. In each test, participants are presented with two 10-second video clips synthesized by different models (including the ground truth) for the same speech, played sequentially. Participants are required to select their preferred clip based on the instruction displayed below the videos and rate their preference on a scale of $0$ to $2$, with $0$ indicating no preference. The other clip in the video pair automatically receives the opposite score (e.g., if the participant rates the preferred video $1$, the unselected video gets a score of $-1$). We recruit participants through the Credamo platform \cite{credamo}.

We conduct four types of preference tests: \emph{human likeness}, \emph{appropriateness}, \emph{style correctness (with dataset label)}, and \emph{style correctness (with random prompt)}. All tests include attention checks. For the \emph{human likeness} test, participants are asked whether the generated motion resembles the motion of a real human. The video clips are muted to prevent any influence from the speech. In the \emph{appropriateness} test, participants rate whether the generated motion matches the rhythm and semantics of the speech. For the \emph{style correctness (w/ dataset label)} test, participants are required to assess how well the generated motion represents the given text style prompt. Note that the text prompt contains a style label provided by the dataset in this test, and the videos are muted. Lastly, a similar test, \emph{style correctness (with random prompt)}, is conducted to evaluate the system's generalizability. The text prompt used in this test does not contain any style label from the dataset, and the motion style that the text prompt describes may not exist in the dataset. We provide more details about the user study in Appendix \ref{sec:details_of_user_study}.

On the BEAT dataset, we only compare our system with CaMN using \emph{human likeness} and \emph{appropriateness} tests, as CaMN does not support text prompt-conditioned style control in the original paper. In practice, we compare four methods: the ground-truth gestures (GT), our system (Ours), our system without transcript input (w/o transcript) for ablation, and CaMN. All the systems take only speech as input, with our system generating motion in the style-unconditional mode (w/o style prompt), and the extra speaker ID input of CaMN defaulting to the ground truth.

In this user study, $100$ and $98$ subjects pass the attention checks for the human likeness and appropriateness tests, respectively. Table \ref{tab:user_study} shows the average scores obtained in these tests. We conducted a one-way ANOVA and a post-hoc Tukey multiple comparison test for each user study. For the \emph{human likeness} test, GT, Ours, and Ours (w/o transcript) are statistically tied and outperform CaMN ($p < 0.001$). As for the \emph{appropriateness} test, Ours receives a higher score compared to other methods ($p < 0.001$), but the score of Ours (w/o transcript) drops significantly due to the lack of semantic information. Based on these two motion quality-related studies, we conclude that our diffusion-based model outperforms CaMN, and the semantic modules (including the semantics-aware attention layer and $\mathcal{L}_{\eqword{semantic}}$) are crucial to ensuring semantic consistency between speech and generated gestures.

On the ZeroEGGS dataset, we evaluate our system using all four preference tests. For the \emph{human likeness} and \emph{appropriateness} tests, we compare our system with ZE. For the \emph{style correctness} tests, we compare our system with MD-ZE. We let our system generate gestures based on motion prompts for a fair comparison, where the motion prompt is randomly sampled from the training set containing four types of style (happy, sad, angry, and old). Two speech recordings in \emph{neutral} style are selected as the test set to prevent the potential style information embedded in the speech from affecting the generated styles. Four text prompts are prepared for the \emph{style correctness (w/ dataset label)} test: $\{$\emph{the person is happy}, \emph{the person is sad}, \emph{the person is angry}, \emph{an old person is gesticulating}$\}$. As for the \emph{style correctness (w/ random prompt)} test, the test style prompts either specify a speaker style (\emph{Hip-hop rapper}), define a pose (\emph{holding a cup with the right hand} and \emph{looking around}), or describe an abstract condition (\emph{a person just lost job}). Note that there is no ground truth for these test style prompts in the ZeroEGGS dataset.

After the attention checks, we recorded the answers of $99$ participants for the \emph{human likeness} and \emph{appropriateness} tests. As shown in Table~\ref{tab:user_study}, the difference between human-likeness scores of Ours and ZE is not significant ($p < 0.05$), while Ours outperforms ZE in the speech-matching metric (appropriateness) by a clear margin ($p < 0.001$). This is because ZE only takes speech audio as input and may lack sufficient semantic information. 
In the \emph{style correctness} tests, $100$ valid subjects score Ours higher compared to MD-ZE for text prompts containing style labels of the dataset ($p < 0.001$). The left part of \fig\ref{fig:comparison_results} offers a visualization demo, where the result of MD-ZE reflects some sense of sadness, but the motion of Ours is more vivid. These results confirm the efficiency of our system in style control. Our method remains robust even when given random prompts, ahead of MD-ZE ($p < 0.001$), and generates accurate stylized gestures. The visualization results are shown in the right part of Figure~\ref{fig:comparison_results}.
\begin{table*}[t]
    \centering
    \caption{Quantitative evaluation on the BEAT and ZeroEGGS datasets. Motion quality-related metrics (FGD, SRGR, and SC) are only calculated on the big BEAT dataset to guarantee the accuracy of approximation of the motion distribution (FGD and SRGR) and the generalizability of the learned semantic space (SC). Style-related metric (SRA) is measured on the ZeroEGGS dataset because it has a variety of styles and the comparison system MD-ZE is trained on it. This table reports the mean ($\pm$ standard deviation) values for each metric by synthesizing on the test data $10$ times.}
    \label{tab:quantitative_evaluation}

    \newcolumntype{Y}{>{\raggedleft\arraybackslash}X}
    \newcolumntype{Z}{>{\centering\arraybackslash}X}
    \begin{tabularx}{\linewidth}{llYYYY}
        \toprule
        Dataset & System & FGD $\downarrow$ & SRGR $\uparrow$ & SC $\uparrow$ & SRA ($\%$) $\uparrow$ \\ 
        \toprule
        \multirow{4}*{BEAT} & GT & - & $1.00$ & $0.80$ & - \\
        \cline{2-6}
        & CaMN & $110.23 \pm 0.00$ & $0.25 \pm 0.00$ & $0.33 \pm 0.00$ & - \\
        & Ours (w/o transcript) & $97.82 \pm 2.56$ & $0.09 \pm 0.02$ & $0.11 \pm 0.03$ & - \\
        & Ours & $\bm{85.17 \pm 3.35}$ & $\bm{0.51 \pm 0.08}$ & $\bm{0.58 \pm 0.15}$ & - \\
        
        \midrule
        \multirow{4}*{ZeroEGGS} & MD-ZE & - & - & - & $47.50 \pm 0.97$ \\
        & Ours (w/o $\mathcal{L}_{\eqword{style}}$) & - & - & - & $64.28 \pm 2.17$ \\
        & Ours (concatenation fusion) & - & - & - & $68.15 \pm 2.02$ \\
        & Ours & - & - & - & $\bm{71.53 \pm 1.01}$ \\
        \bottomrule
    \end{tabularx}
\end{table*}

\subsection{Quantitative Evaluation}
\label{subsec:quantitative_evaluation}
We quantitatively measure the motion quality, speech-gesture content matching, and style correctness using three metrics : Fr{\'e}chet Gesture Distance (FGD) \cite{yoon2020trimodalgesture}, Semantics-Relevant Gesture Recall (SRGR) \cite{liu2021beatdataset}, Semantic Score (SC), and Style Recoginition Accuracy (SRA) \cite{jang2022motionpuzzle}. 

The FGD measures the distance between the distributions of latent features calculated from generated and real gestures, respectively. This metric is typically used to assess the perceptual quality of gestures. A lower FGD usually suggests higher motion quality. 

It has been shown that vanilla L1 distance and Probability of Correct Keypoint (PCK) are not suitable for assessing gesture performance due to the inherent many-to-many correspondence between speech and gestures \cite{yoon2020trimodalgesture,liu2021beatdataset}. To address this issue, we adopt the SRGR metric proposed by \citet{liu2021beatdataset}, which uses manually-labeled semantic scores as the weights for PCK. This metric ensures that semantically relevant gestures align closely with the ground truth, while allowing for variation in other gestures, such as beat gestures. Besides, SRGR partially captures the diversity of generated gestures according to \cite{liu2021beatdataset}. A higher SRGR indicates better performance.

To evaluate the semantic coherence between speech and generated gestures, we propose a new metric, namely the semantic score (SC), to calculate the semantic similarity between generated motion and the ground-truth transcripts in the gesture-transcript embedding space (Section \ref{sec:gesture-transcript_embedding_learning}).  SC can be computed as
\begin{align}
    \eqword{SC} = \cos(\vect{z}^{t}, \vect{z}^{g*}_0),
\end{align}
where $\vect{z}^{t}$ and $\vect{z}^{g*}$ are the ground-truth transcript encoding and the gesture encoding of the generated motion, respectively. SC $\in [-1, 1]$ and a higher SC suggests better speech-gesture content matching.

Following \cite{jang2022motionpuzzle}, we pre-train a classifier on the ZeroEGGS dataset to predict motion style labels. The dataset contains 19 distinct styles, but we exclude some ambiguous ones, such as \emph{agreement}, \emph{disagreement}, \emph{oration}, \emph{pensive}, \emph{sarcastic}, and \emph{threatening}. For testing, we use all speech recordings in the \emph{neutral} style as input speech. Text prompts are generated using a prompt template (\emph{the person is [style label]}), where the style label (e.g., happy and sad) is from the dataset.  We then use the neutral speech and synthetic text prompts as inputs to generate stylized gestures. Lastly, we employ the style classifier to measure the style recognition accuracy (SRA) on the test set. A higher SRA indicates better performance in terms of style control. It is important to note that SRA is limited to covering styles that appear in the dataset.

We conduct motion quality-related evaluations, specifically FGD, SRGR, and SC, on the BEAT dataset only, as a large dataset is necessary to ensure accurate approximation of the motion distribution (FGD and SRGR) and generalizability of the learned semantic space (SC). For style-related evaluations (SRA), we measure performance on the ZeroEGGS dataset only, as it has a variety of styles and the baseline system MD-ZE is trained on it. The FGD, SRGR and SC are calculated using sentence-level motion segments, while 10-second segments are used to measure the SRA. We compute the mean ($\pm$ standard deviation) values for each metric by synthesizing on the test data 10 times. The test set of the BEAT dataset is composed of the first file of both \emph{conversation} and \emph{self-talk} sessions for each speaker who appears in the training set.

As shown in Table \ref{tab:quantitative_evaluation}, our system outperforms the baseline CaMN in motion quality-related metrics (FGD, SRGR and SC). The SC value of our system significantly decreases when the transcript input is discarded, highlighting the importance of the gesture-transcript embedding module. In terms of style control, our system outperforms the baseline MD-ZE in the SRA metric by a clear margin, which is consistent with the results of the user study (see Section \ref{subsubsec:user_study}). The SRA results of different ablation settings, i.e., Ours (w/o $\mathcal{L}_{\eqword{style}}$) and Ours (concatenation fusion), demonstrate the necessity of these components in our system.

\subsection{Application}
\label{subsec:application}
Our system enables several interesting applications. One such application involves enhancing co-speech gestures by specifying the style of each sentence in the speech and using style prompts to guide the performance of the character. This process can be automated by a large language model, such as ChatGPT \cite{openai2022chatgpt}. Specifically, we can instruct ChatGPT to generate a style prompt for each sentence in the speech and then generate stylized co-speech gestures accordingly. \fig\ref{fig:application_style_prompt_generation} demonstrates the editing results (the first row) obtained by using generated text prompts, which are more vivid than the style-unconditional results (the second row).

\begin{figure*}[t]
    \centering
    \includegraphics[width=\textwidth]{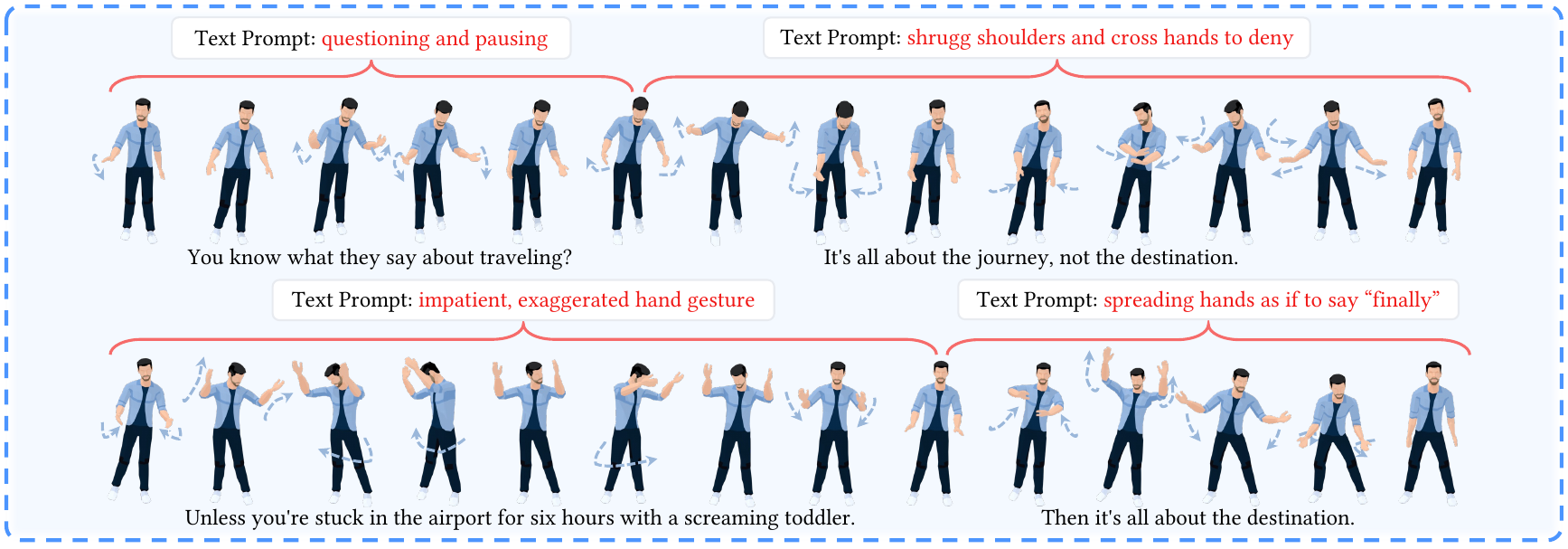}
    \caption{Qualitative result of gesture editing (Section \ref{subsec:application}). Both the speech transcript and text prompts are generated by ChatGPT \cite{openai2022chatgpt}. Additionally, we translate the generated transcript into audio using a Text-To-Speech tool \cite{murfai2022tts} and use the resulting audio as the speech input. Note that such synthesized voices are not seen during training.}
    \label{fig:application_text_generation}
    \Description{}
\end{figure*}

Another application is to let ChatGPT write a story and creating the corresponding style prompts. Then, we can translate the generated transcript into audio using a Text-To-Speech tool \cite{murfai2022tts} and use the audio as the speech input. This can result in a skillful storyteller. As shown in \fig\ref{fig:application_text_generation}, we let ChatGPT write a short joke about \emph{travel} with suitable text prompts and synthesize gestures using them. The generated gestures successfully portray the joke with diverse body styles.

The full prompt inputs to ChatGPT of these application examples are detailed in Appendix \ref{sec:prompts_for_chatgpt}. Please refer to the supplementary video for animation results.

\subsection{Ablation Study}
We analyze the impact of the gesture-transcript embedding, the style loss, and the style fusion mechanism on our performance. The results are reflected in Table \ref{tab:user_study}, Table \ref{tab:quantitative_evaluation}, and the supplementary video.

\subsubsection{Gesture-Transcript Embedding}
In this experiment, we omitted the transcript embedding input of the denoising network, replaced the semantics-aware attention layer with the causal self-attention layer, and retrained the generator without the semantic loss $\mathcal{L}_{\eqword{semantic}}$. The supplementary video demonstrates that the generated motion maintains rhythmic harmony but lacks reasonable semantics. Moreover, semantics-aware metrics, such as appropriateness (Table \ref{tab:user_study}) and SC (Table \ref{tab:quantitative_evaluation}), also show a significant drop for our model (w/o transcript). These results confirm that the gesture-transcript embedding module effectively enhances the semantic consistency between speech and gestures.

\subsubsection{Style Loss}
In this experiment, we retrain the denoising network without the style loss $\mathcal{L}_{\eqword{style}}$. The supplementary video demonstrates that the recognizability of the generated style is reduced. The style recognition accuracy (SRA) in Table \ref{tab:quantitative_evaluation} also decreases. An intuitive explanation is that the style-relevant \emph{knowledge} embedded in CLIP serves as a guide for the stylization of the generated gestures through the style loss. This guidance approximates style-aware supervision, even in the absence of explicit labels.

\subsubsection{Style Fusion Mechanism}
In this experiment, we replace the AdaIN-based style embedding fusion scheme with direct concatenation, where the style embedding is broadcasted and concatenated to the intermediate deep features in the generator for style modification. Although the SRA value of Ours (concatenation fusion) drops only  slightly compared to AdaIN, the supplementary video shows that the motion generated by Ours (concatenation) exhibits jittering and unnatural movements. 
\section{Conclusion}
\label{sec:conclusion}
In this paper, we have presented GestureDiffuCLIP, a CLIP-guided co-speech gesture synthesis system that generates stylized gestures based on arbitrary style prompts while ensuring semantic and rhythmic harmony with speech. We leverage powerful CLIP-based encoders to extract style embeddings from style prompts and incorporate them into a diffusion model-based generator through an AdaIN layer. This architecture effectively guides the style of the generated gestures. The CLIP latents make our system highly flexible, supporting short texts, motion sequences, and video clips as style prompts. We also develop a semantics-aware mechanism to ensure semantic consistency between speech and generated gestures, where a joint embedding space is learned between gestures and speech transcripts using contrastive learning. Our system can be extended to achieve style control of individual body parts through noise combination. We conduct an extensive set of experiments to evaluate our framework. Our system outperforms all baselines both qualitatively and quantitatively, as evidenced by FGD, SRGR, SC, and SRA metrics, and user study results. Regarding application, we demonstrate that our system can effectively enhance co-speech gestures by specifying style prompts for each speech sentence and using these prompts to guide the character's performance. We can further automate this process by employing a large language model like ChatGPT, enabling a
skillful storyteller.
\begin{figure*}[t]
    \centering
    \includegraphics[width=\linewidth]{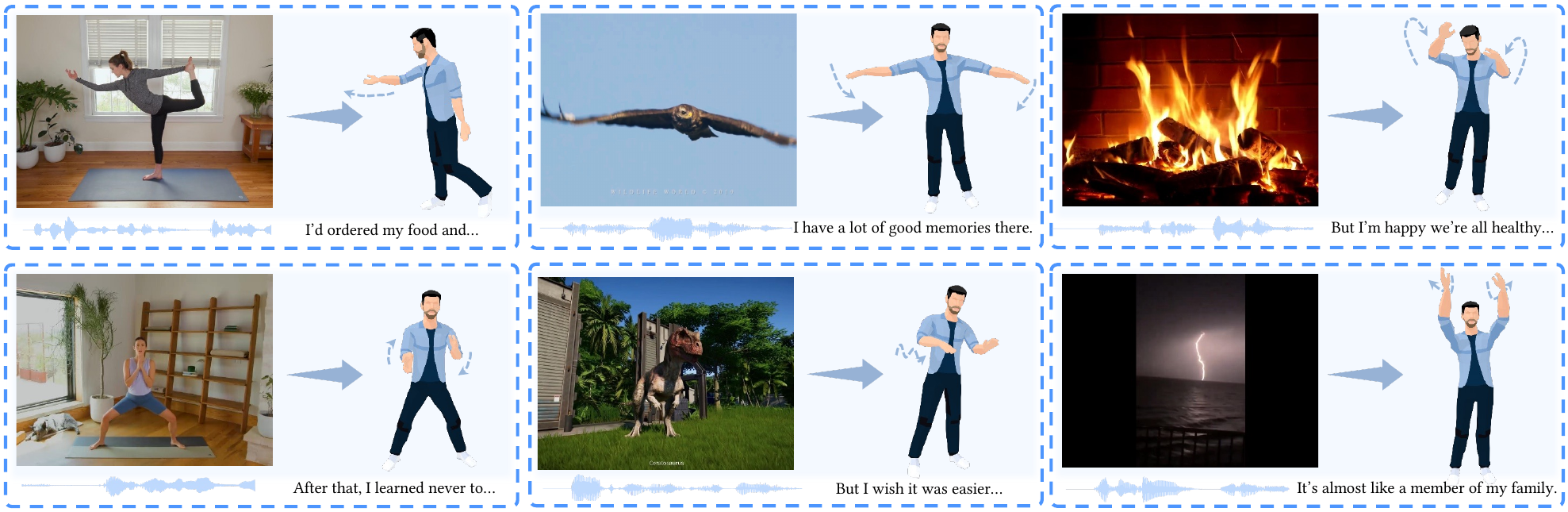}
    \caption{Gestures synthesized by our system conditioned on video prompts: Yoga poses \cite{adriene2021yoga1video,adriene2021yoga2video}, flying bird \cite{wildlife2019birdvideo}, dinosaur \cite{dinosaur2018dinosaurvideo}, fire \cite{fire2016firevideo}, and lightning \cite{manabouttown12021lightning}.}
    \Description{}
    \label{fig:video_prompt_result}
\end{figure*}

In our system, styles refer to the overall appearance of motion or, more precisely, the aspects of gestures that are independent of input speech content. This encompasses both stylistic features and motion constraints. Our system processes these two types of style within a unified framework, potentially enabling users to define them using a combined prompt, such as \emph{the person is happy and holds a cup of coffee in their right hand} (stylistic feature + constraints), when a powerful language model is available. Another interesting extension to our current framework could be to allow users to combine different forms of prompts for a more accurate description of a style, such as specifying a stylistic feature with a text prompt and defining motion details using a video or motion prompt. This presents a compelling direction for future work.

Our system achieves zero-shot style control using unseen style prompts and generalizes to some unseen voices, such as the one synthesized by the Text-to-Speech tool used in \fig\ref{fig:application_text_generation}. However, as is common with deep learning-based approaches, the capacity and robustness of our system are constrained by the training data and the network architecture. For instance, due to the limitations of the vanilla CLIP text encoders, our system cannot accept excessively long text prompts, and users may find that some descriptions are not interpreted accurately. When employing motion clips or human videos as style prompts, our system consistently endeavors to generate gestures with similar poses to those displayed in the prompt. Nevertheless, as demonstrated in Figure \ref{fig:video_prompt_result}, poses that diverge significantly from the dataset, such as many yoga poses, may not be accurately reproduced. Furthermore, since non-human videos typically lack a lucid motion semantic interpretation, the semantic connection between the video content and the generated motion style can be ambiguous. As illustrated in Figure \ref{fig:video_prompt_result}, our video encoder extracts arm poses from the bird video and the dinosaur video but interprets the lightning video as the \emph{raise both hands} prompt, possibly due to the shape of the lightning and the clouds. Lastly, audio inputs with acoustic features that differ substantially from the training dataset can result in unsatisfactory outcomes, where the generated motion may fail to adhere to the speech rhythm or correspond with the speech content. Enhancing the system's generalizability and interpretability remains a practical challenge for further exploration.

We learn the gesture-transcript joint embedding space using a CLIP-style contrastive learning framework, which has been shown to be effective in extracting semantic information from both modalities and enabling applications such as motion-based text retrieval and word saliency identification. CLIP-style models typically scale well and have the potential for increased power when trained on larger datasets \cite{radford2021clip}. Investigating this potential presents an intriguing avenue for future research.

The generated motion of our system exhibits slight foot sliding, which is a common problem in kinematically-based methods \cite{holden2017pfnn,tevet2022humanmotiondiffusion}. This can be alleviated via IK-based post-processing, but unnatural transitions in contact status may arise after post-processing \cite{li2022ganimator}. Developing a physically-based co-speech gesture generation system could fundamentally address this problem.

Finally, our system is based on a latent diffusion model, which effectively ensures motion quality but requires a large number of diffusion steps during inference, making it challenging to synthesize motions in real-time. Acceleration techniques, such as PNDM \cite{liu2022pseudo}, could be considered in the future to optimize the sampling efficiency.

\begin{acks}
  We would like to thank the anonymous reviewers for their constructive suggestions and feedback. We also thank Baoquan Chen for various discussions and help. This work was supported in part by start-up grants from Peking University.
\end{acks}

\bibliographystyle{ACM-Reference-Format}
\bibliography{gesture}

\clearpage

\appendix

\section{Details of User Study}
\label{sec:details_of_user_study}
A comparison pair contains two $10$-second videos that are played in order from left to right. We generate each pair using the same speech and character model. The user study questionnaires are created using the Human Behavior Online (HBO) tool offered by the Credamo platform \cite{credamo}. This tool is designed to conduct psychological experiment sample collection without complex programming. All tests and questionnaires are composed of $24$ video pairs. An experiment takes an average of $10$ minutes to complete. We recruit participants from the US and China through Credamo. Participants who take tests with sound are required to speak English fluently. Following \cite{alexanderson2022diffusiongesture}, an attention check is randomly introduced in the experiment to screen valid samples. Specifically, a text message: \emph{attention: please select the rightmost option} appears both at the bottom of the video pair for the whole duration of the question and in the video during the transition gap between the two clips. Samples that fail the attention check are not used for final results.

\subsection{Motion-Quality Study}

\subsubsection{BEAT Dataset}
We select 24 speech segments from the BEAT dataset's test set to create gestures, generating 24 video clips for each method. We compare four approaches: GT, Ours, Ours (w/o transcript), and CaMN, leading to 12 potential pairwise combinations for side-by-side demonstrations. This results in a total of 288 video pairs (24 speech samples $\times$ 12 combinations). Each participant is asked to assess 24 video pairs, encompassing all 24 speech samples. Each of the 12 possible comparison combinations appears twice. The speech samples and pairing of comparisons are randomized for each participant.

\subsubsection{ZeroEGGS Dataset}
\label{subsubsec:motion-quality_study_ZeroEGGS}
We select $6$ audio clips from the ZeroEGGS test recordings in neutral style (\emph{003\_Neutral\_2} and \emph{004\_Neutral\_3}) to synthesize gestures based on $4$ different styles (\emph{happy}, \emph{sad}, \emph{angry}, and \emph{old}), yielding $24$ video clips for each system. We compare two systems, ZE and Ours, in this study. During the assessment, each participant encounters all 24 video clips ($6$ audio clips $\times$ $4$ styles) once in a randomized order, with the outcomes of ZE and Ours evenly distributed in the front position of each video pair.

\subsection{Style-Control Study on the ZeroEGGS Dataset}
The configurations for the \emph{style correctness (w/ dataset label)} and \emph{style correctness (w/ random prompt)} tests are identical, except for the input text prompt. 
In each study, we use the same $6$ audio clips from experiment \ref{subsubsec:motion-quality_study_ZeroEGGS} to generate motions conditioned on $4$ text prompts. This leads to 24 video clips for each system, MD-ZE and Ours, in each test. 
For the \emph{style correctness (w/ dataset label)}, the text prompts are: $\{$\emph{the person is happy}, \emph{the person is sad}, \emph{the person is angry}, \emph{an old person is gesticulating}$\}$. For the \emph{style correctness (w/ random prompt)} test, the text prompts are: $\{$\emph{Hip-hop rapper}, \emph{holding a cup with the right hand}, \emph{looking around}, \emph{a person just lost job}$\}$.
Again, each participant evaluates these $24\times{}2$ video clips in pairs in a randomized sequentially, where the resulting motion of MD-ZE and Ours evenly distributed in the front position of each video pair.

\section{Implementation Details of Baselines}
\label{sec:implementation_details_of_baselines}
\begin{figure}[t]
    \centering
    \includegraphics[width=\linewidth]{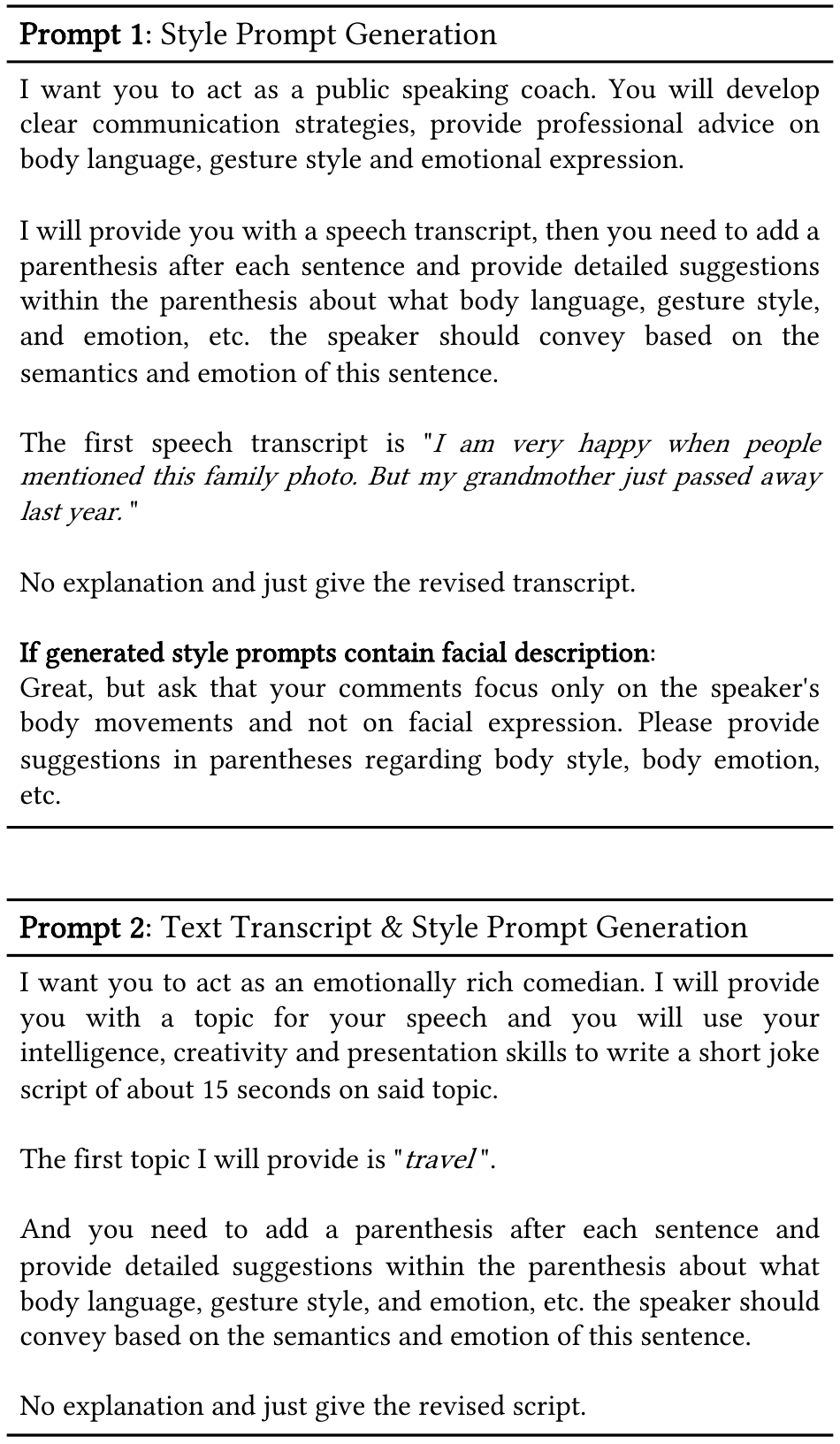}
    \caption{Prompt inputs to ChatGPT \cite{openai2022chatgpt}.}
    \Description{}
    \label{fig:chatGPT_prompts}
\end{figure}
At the time of writing this work, the authors of CaMN \cite{liu2021beatdataset} have not provided the pre-trained generation model. Instead, they offered training codes for a toy dataset and a pre-trained motion auto-encoder for the calculation of FGD. We run the provided training codes on a larger dataset used in the original paper, discarding the unreleased emotion label from the conditions of the model. The FGD value of the reproduced model is $122.5$, which is close to the value reported in the original paper ($123.7$). The visual quality of gestures synthesized by the reproduced model is similar to that shown in the video demo of CaMN. We then follow the configuration above and train a new CaMN model on a part of the BEAT dataset used in our work (Section \ref{subsec:system_setup}). This model achieves better performance on FGD ($110.23$) and is utilized as the baseline in this paper.

For the baseline MD-ZE on the ZeroEggs dataset, the two components of this baseline, i.e., MotionDiffuse (MD) \cite{zhang2022motiondiffuse} and ZeroEGGS (ZE) \cite{ghorbani2022zeroeggs}, are constructed using the official pre-trained models. Note that the skeletons of the two models are different. We thus retarget the motion prompt generated by MD to fit the interface of ZE. Specifically, we first convert the generated motion prompt, represented as SMPL \cite{loper2015smpl} joint positions, into joint rotations and save them as a BVH file. Then, we retarget the prompt in SMPL skeleton to the ZeroEGGS skeleton using a Blender add-on, \emph{BVH Retargeter} \cite{padovani2020bvhretargeter}.

\section{Prompts for ChatGPT}
\label{sec:prompts_for_chatgpt}
\fig\ref{fig:chatGPT_prompts} demonstrates the prompt inputs for ChatGPT \cite{openai2022chatgpt} in Section \ref{subsec:application}.

\end{document}